# Ampe-DB: A Dataset for Human Activity and Interaction Analysis Using the Ghanaian Game Ampe

Authors : Ruth Senior Biney & Isaac Kojo Yedonu Aboah

## Abstract

Human activity recognition has benefited substantially from large-scale video datasets, yet many culturally situated forms of movement and interaction remain underrepresented in existing benchmarks. We introduce **Ampe-DB**, a multimodal dataset for human activity and interaction analysis based on *Ampe*, a traditional Ghanaian game built around coordinated jumping, leg movements, and pairwise interaction. Ampe-DB contains **637 canonical interaction clips involving 15 participants**, with RGB video, pose representations, participant and pair metadata, four-class movement labels, and binary movement-relation labels.

We define two supervised benchmark tasks: **four-class movement recognition** and **binary movement-relation classification**. To distinguish conventional recognition performance from generalization across people and interactions, we evaluate models under three complementary protocols: a duplicate-safe fixed benchmark, a four-fold pair-held-out benchmark, and participant-held-out diagnostic evaluation for participants with complete target-class coverage. Baselines include a frozen Kinetics-400-pretrained R(2+1)D-18 RGB representation, a two-layer bidirectional GRU operating on two-person pose sequences, and validation-selected RGB–pose late fusion.

All learned systems achieve strong performance on the duplicate-safe fixed benchmark. However, substantial differences emerge under held-out evaluation. For four-class recognition, RGB macro-F1 decreases from **0.955** on the fixed benchmark to **0.418** under pair-held-out evaluation and **0.384** under participant-held-out evaluation. Pose representations remain substantially more robust, achieving **0.957**, **0.811**, and **0.839** macro-F1 across the same protocols. Late fusion achieves the strongest fixed and four-class pair-held-out results but does not consistently outperform pose under participant shift. Binary movement-relation classification likewise shows marked generalization difficulty, particularly for different-leg configurations.

These results demonstrate that high fixed-split performance can substantially underestimate the difficulty of generalizing Ampe recognition across unfamiliar interactions and participants. Ampe-DB provides a culturally grounded and technically challenging benchmark for fine-grained movement recognition, human-human interaction analysis, multimodal representation learning, and identity-aware generalization.

# 1. Introduction

Human activity recognition is a longstanding problem in computer vision, with applications in human-computer interaction, assistive technologies, sports analysis, robotics, and behavioural understanding. Progress in the field has been closely linked to benchmark datasets that enable models to learn spatial and temporal patterns of human movement and provide standardized settings for comparing representations. Large-scale resources such as Kinetics have supported the development of video-based action-recognition models, while datasets such as NTU RGB+D have enabled systematic evaluation using RGB, depth, and skeletal representations across different subjects and viewpoints (Carreira and Zisserman, 2017; Shahroudy et al., 2016).

Despite this progress, the activities represented in widely used benchmarks capture only part of the diversity of human movement and interaction. Many culturally situated forms of play, performance, and social interaction remain largely absent from computational datasets. Expanding this coverage is important not only for geographic and cultural representation, but also because different activities can introduce distinct technical problems involving coordination, fine-grained movement, temporal structure, and relationships between multiple people.

In this work, we study these questions through Ampe, a traditional Ghanaian game based on rapid coordinated movement between two players (Coker, 2022). During an exchange, both participants jump and extend a leg, producing a movement configuration determined by the leg selected by each player. The interaction is represented through four movement configurations: left_left, left_right, right_left, and right_right. These configurations can also be grouped according to whether the two movement tokens form a same-leg or different-leg configuration.

Although the physical movement is compact, Ampe presents a non-trivial recognition problem. The distinction between classes can depend on relatively small changes in leg position, and the complete interaction state cannot be determined from one participant alone. The movement must be interpreted relationally across two simultaneously acting people. Moreover, repeated recordings of the same individuals and participant pairs create a risk that models may exploit visual information associated with familiar people or interaction contexts instead of learning movement representations that transfer across performers.

This issue makes evaluation design particularly important. A model can perform strongly on a conventional train-test split even when its learned representation transfers poorly to unfamiliar people or pairings. Exact duplicate removal addresses one source of leakage but does not by itself test generalization across participant identity or interaction structure. Ampe therefore provides a useful setting in which to distinguish recognition under a relatively familiar test distribution from recognition when interaction pairs or participants are held out during model development.

We introduce **Ampe-DB: A Dataset for Human Activity and Interaction Analysis Using the Ghanaian Game Ampe**, a multimodal dataset containing **637 canonical two-person interaction clips involving 15 participants**. Ampe-DB provides RGB video, two-person pose representations, participant and pair metadata, four-class movement annotations, and binary movement-relation labels. The audited canonical dataset contains **557 unique audiovisual SHA-256 hashes**, allowing exact duplicate content to be explicitly identified and controlled during evaluation.

We formulate two supervised benchmark tasks: **four-class movement recognition** and **binary movement-relation classification**. To evaluate different forms of generalization, we establish three complementary protocols: a duplicate-safe fixed benchmark, a four-fold pair-held-out benchmark, and participant-held-out diagnostic evaluation for participants with complete target-class coverage. The pair-held-out protocol serves as the principal generalization benchmark, while the participant-held-out setting provides a complementary diagnostic of identity transfer.

We further provide reproducible baselines using a frozen Kinetics-400-pretrained R(2+1)D-18 RGB representation, a bidirectional gated recurrent unit model operating on two-person pose sequences, and validation-selected RGB–pose late fusion. The experiments show that the choice of evaluation protocol substantially changes the apparent difficulty of the task. RGB, pose, and fusion models all achieve high performance on the duplicate-safe fixed benchmark, but their behaviour diverges under held-out evaluation. Pose representations remain substantially more robust than frozen RGB features for four-class recognition across unseen participant pairings and unseen participants, while simple late fusion does not provide consistent gains across all settings.

These findings position Ampe-DB not simply as a collection of videos from an underrepresented cultural activity, but as a benchmark for studying how models represent and generalize fine-grained human interaction.

The principal contributions of this work are:

1. **A multimodal dataset centred on a Ghanaian human interaction.** We introduce Ampe-DB, comprising 637 canonical Ampe interaction clips from 15 participants, with RGB video, two-person pose representations, participant and pair metadata, fine-grained movement labels, and derived movement-relation labels.
2. **A structured formulation of movement and interaction recognition.** We define a four-class task that preserves the stored two-token movement configuration and a binary task that captures whether the two movement tokens form a same-leg or different-leg relationship.
3. **Leakage-aware protocols for evaluating generalization.** We provide a duplicate-safe fixed benchmark, four-fold pair-held-out evaluation, and participant-held-out diagnostic protocols designed to separate conventional recognition performance from transfer across unfamiliar interactions and identities.

4. **Reproducible RGB, pose, and multimodal reference baselines.** Across three predefined training seeds, the experiments reveal a substantial generalization gap for frozen RGB features, stronger robustness from pose representations, and limitations of simple validation-selected late fusion under participant shift.

Together, these contributions establish Ampe-DB as a culturally grounded benchmark for fine-grained human activity recognition, human-human interaction analysis, multimodal learning, and generalization across people and interaction contexts.

# 2. Related Work

## 2.1 Human Activity Recognition Datasets

Progress in human activity recognition has been closely associated with the development of increasingly large and diverse benchmark datasets. Kinetics established a large-scale video benchmark containing hundreds of human-action categories collected from unconstrained Internet video and became an important source of pretraining for video-recognition models (Carreira and Zisserman, 2017). NTU RGB+D introduced more than 56,000 activity samples from 40 participants and combined RGB, depth, infrared, and skeletal information with cross-subject and cross-view evaluation protocols (Shahroudy et al., 2016). Its later extension, NTU RGB+D 120, expanded the benchmark to more than 114,000 samples, 120 activity classes, and 106 participants (Liu et al., 2020).

Other datasets have focused on more specific dimensions of human activity understanding. HAA500 emphasizes fine-grained atomic actions, with 500 categories curated around precise human movements (Chung et al., 2021). Ego4D instead addresses large-scale egocentric activity understanding, providing thousands of hours of first-person video across multiple countries and supporting tasks spanning activity, social interaction, forecasting, and episodic memory (Grauman et al., 2022).

These datasets demonstrate that the value of an activity benchmark is determined not only by its scale but also by the type of behaviour it captures and the research questions enabled by its annotations and evaluation protocol. Ampe-DB follows this task-focused approach. Rather than attempting to represent a broad vocabulary of unrelated activities, it concentrates on a single two-person game in which a small number of closely related body configurations must be distinguished. The benchmark therefore emphasizes fine-grained movement, relational interpretation, and transfer across participants and interaction pairs.

## 2.2 Video-Based Action Representations

Video action recognition requires representations that capture both visual appearance and temporal change. Early deep-learning approaches such as two-stream convolutional networks explicitly separated spatial appearance information from temporal motion information before combining their predictions (Simonyan and Zisserman, 2014).

Subsequent architectures increasingly learned spatial and temporal information jointly. R(2+1)D factorizes three-dimensional convolutions into separate spatial and temporal components, enabling the network to model video dynamics while retaining the advantages of residual convolutional architectures (Tran et al., 2018).

This architecture is relevant to Ampe-DB because each interaction unfolds across a short temporal sequence. The discriminative movement is not limited to a static scene: participants prepare, jump, extend their legs, and return toward the ground over successive frames.

At the same time, RGB video contains substantially more information than the movement labels require. A video representation may encode clothing, body appearance, background, illumination, camera characteristics, and other contextual information alongside the action itself. In datasets containing repeated performers, some of these signals may correlate with classes in the development data without transferring reliably to unfamiliar participants or interactions.

Ampe-DB therefore evaluates the RGB representation under both conventional duplicate-safe evaluation and protocols that explicitly hold out participant pairings or individual participants. The objective is not to assume that RGB representations rely on identity-related shortcuts, but to measure whether the representation learned under one interaction distribution transfers when that distribution changes.

### 2.3 Pose and Skeleton-Based Activity Recognition

Pose-based activity recognition represents human behaviour through the locations and dynamics of body landmarks. Compared with raw RGB video, this representation places greater emphasis on body geometry and motion while discarding much of the surrounding image appearance.

Skeleton information has played an important role in multimodal activity datasets such as NTU RGB+D (Shahroudy et al., 2016). Approaches to skeleton-based recognition have included recurrent networks, convolutional models, and graph-based architectures. Spatial-Temporal Graph Convolutional Networks, for example, represent body joints as graph nodes and learn spatial and temporal dependencies directly from skeletal sequences (Yan et al., 2018).

Pose is particularly relevant to Ampe because the benchmark classes are defined by the body configurations of two interacting players. The difference between left_left, left_right, right_left, and right_right is fundamentally related to the relative configuration of their legs.

Ampe-DB therefore provides a useful setting for comparing an appearance-rich RGB representation with a representation more directly centred on body movement. Baseline V1 uses a bidirectional gated recurrent unit model rather than a specialized skeleton architecture. This intentionally simple model establishes a reference point from which future graph-based, transformer-based, or interaction-specific pose models can be evaluated.

### 2.4 Human-Human Interaction Understanding

Many activity-recognition problems involve more than one person. In such settings, the meaning of an action may depend on another person's movement, position, or response rather than on the motion of one individual in isolation.

NTU RGB+D includes mutual actions between participants in addition to individual activities (Shahroudy et al., 2016). More recent datasets have made interpersonal relationships an explicit part of the recognition problem. JRDB-Act, for example, includes annotations for individual actions, interaction-based actions, and social-group activities in crowded real-world environments (Ehsanpour et al., 2022). Ego4D likewise includes social interaction among the behaviours studied from an egocentric perspective (Grauman et al., 2022).

Ampe-DB differs from these broader interaction datasets in the way the interaction itself defines the target. A left_right configuration cannot be determined from the movement of one participant alone because the target describes a joint movement configuration involving both participants. The binary task goes one step further by discarding the specific stored left-right token ordering and retaining only whether the two movement tokens form a same-leg or different-leg relationship. Ampe-DB therefore provides two related levels of interaction analysis: a fine-grained four-class movement state and a coarser relational state derived from the same exchange.

### 2.5 African and Culturally Situated Computer-Vision Data

The availability of locally relevant data remains an important issue in African computer-vision research. Omotayo et al. (2024), in a large-scale review of computer-vision research in Africa, identify access to datasets, computational resources, and research collaboration among the structural constraints affecting the field. Their study also catalogues more than 100 datasets associated with African computer-vision research, illustrating both the growth of locally produced resources and the breadth of domains in which further dataset development remains possible.

For human activity recognition, geographic representation alone does not capture the full question. An activity filmed in an African location may still represent a behavioural category already extensively covered by existing benchmarks. Culturally situated dataset development can instead begin with activities whose structure, meaning, or interaction patterns arise from the local context itself.

Ampe-DB takes this approach by treating the Ghanaian game Ampe as the activity of interest rather than using Ghana only as a location for collecting generic actions.

The resulting computational problem follows directly from the game. Ampe involves two simultaneously acting participants, fine-grained left-right leg configurations, temporal coordination, relational same-leg and different-leg states, and interpretation of one participant's movement in relation to the other.

Cultural context therefore provides the basis for the benchmark's technical structure rather than functioning solely as descriptive metadata.

At the same time, a machine-learning dataset cannot capture the complete cultural meaning or variation of a living practice. Ampe-DB represents the recorded interactions and benchmark formulation described in this paper and should not be interpreted as a comprehensive representation of every form in which Ampe is played or understood.

## 2.6 Positioning of Ampe-DB

Existing work establishes several research directions relevant to Ampe-DB: large-scale video action recognition, fine-grained movement classification, skeleton-based temporal modelling, human-human interaction understanding, and the development of more locally relevant computer-vision datasets.

Ampe-DB sits at the intersection of these areas but is intentionally different in scale and purpose.

It does not seek to compete with Kinetics, NTU RGB+D, Ego4D, or similar benchmarks in the number of videos or action categories. Instead, it provides a focused two-person activity in which closely related movements can be evaluated under explicit controls for exact duplicate content, participant pairings, and participant identity.

Three characteristics distinguish its benchmark formulation.

First, the target categories arise from joint body configurations involving two people, making relational information central to recognition.

Second, the dataset supports both a fine-grained four-class formulation and a binary relational formulation derived from the same movements.

Third, evaluation explicitly separates performance on a duplicate-safe fixed benchmark from generalization to unseen participant pairings and unseen participants.

Ampe-DB is therefore positioned as a compact but challenging benchmark for studying whether models learn transferable structure from human movement and interaction rather than relying only on patterns that remain stable within a familiar evaluation distribution.

# 3. The Game of Ampe and Benchmark Representation

## 3.1 Ampe as a Two-Person Rhythmic Game

Ampe is a traditional Ghanaian game based on rapid, coordinated interaction between two players (Coker, 2022). During a typical exchange, the players face one another and perform a short jump-and-clap movement before independently selecting a leg position as they land. The relationship between the two selected legs contributes to the interpretation of the exchange according to the rule convention being used.

Ampe is therefore both temporal and relational. The relevant movement develops across a short sequence rather than a single static frame, and the complete interaction state depends on the actions of both participants.

For Ampe-DB, the game is represented at the level of individual two-person interaction clips. Each clip captures one movement exchange and is associated with one of four stored two-token movement configurations.

The benchmark abstraction is intentionally narrower than the cultural practice itself. Ampe may vary in playing style, rhythm, terminology, and rule convention across communities. Ampe-DB therefore represents the interactions and movement labels used in this dataset rather than claiming to encode every form of the game.

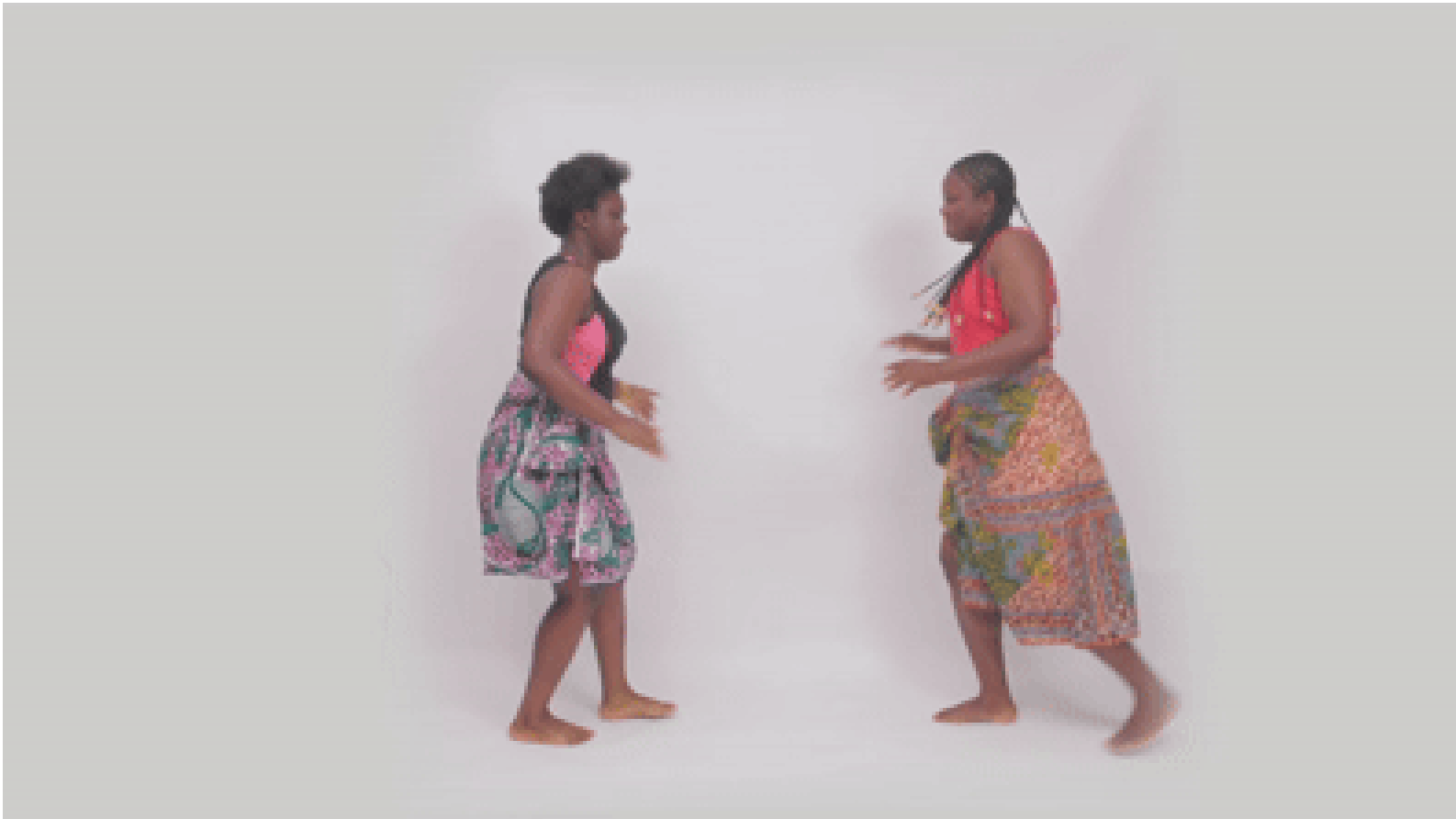

**Figure 1. Representative Ampe interaction sequence.** A short temporal sequence showing two participants progressing through preparation, jump/clap, leg selection, and landing.

### 3.2 Stored Movement Configurations

The primary semantic annotation in Ampe-DB describes a stored two-token movement configuration.

Let $t_1 \in \{L,R\}$ denote the first stored movement token and $t_2 \in \{L,R\}$ denote the second stored movement token. The movement state is therefore:

$$M = (t_1,t_2).$$

This produces four possible movement configurations:

$$\{left_left, left_right, right_left, right_right\}.$$

The labels have the following interpretation:

- left_left: both stored movement tokens have value left;
- left_right: the stored ordering is left, then right;
- right_left: the stored ordering is right, then left; and
- right_right: both stored movement tokens have value right.

The stored token order distinguishes left_right from right_left, and the two are therefore treated as separate classification targets. However, the V1 audit does not independently establish that the first and second movement tokens correspond to the canonical participant fields `player_1_id` and `player_2_id`. The four-class movement configuration itself is verified, but participant-specific meaning should not be assigned to token position without additional provenance.

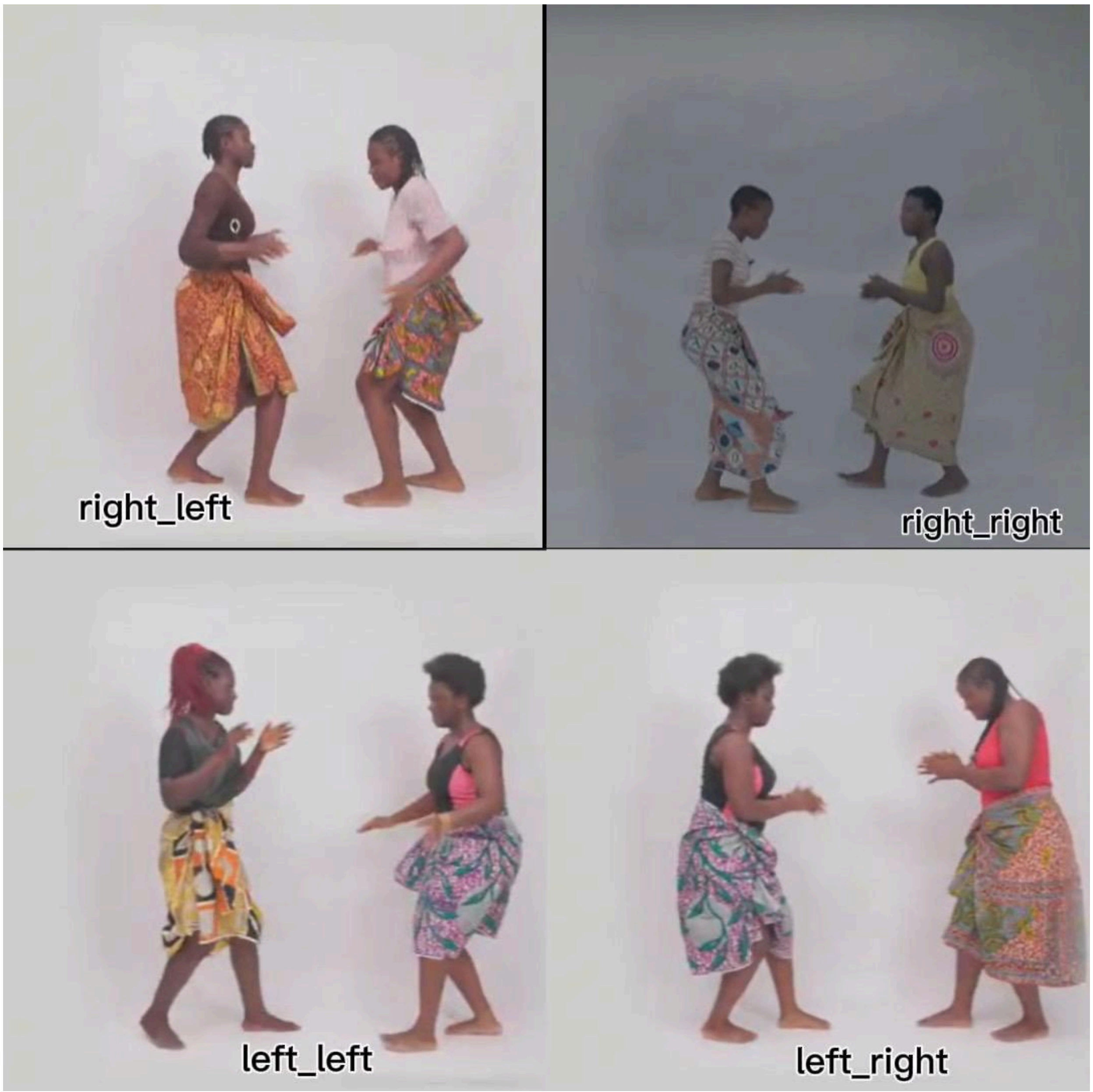

**Figure 2. The four Ampe-DB movement configurations.** Representative RGB and pose examples for left_left, left_right, right_left, and right_right.

### 3.3 Binary Movement Relation

The same four configurations can also be represented through a binary relationship between the two stored movement tokens.

If both movement tokens have the same leg value, the interaction is labelled `same_leg_configuration`:

$$\{left_left, right_right\} \rightarrow same_leg_configuration.$$

If the two movement tokens have different leg values, the interaction is labelled `different_leg_configuration`:

$$\{left_right, right_left\} \rightarrow different_leg_configuration.$$

This binary representation removes the specific left/right token ordering while preserving the relational structure of the movement configuration.

The final V1 audit identifies movement configuration as the verified semantic ground-truth family. The binary movement relation used by the benchmark is therefore derived deterministically from the verified four-class configuration rather than treated as a separately annotated ground-truth source.

### 3.4 Rule-Derived Outcome

Baseline V1 additionally reports an auxiliary rule-derived outcome measure.

For the benchmark implementation, the verified movement relation is mapped deterministically to an encoded player-role outcome. Under this benchmark convention, same-leg and different-leg relations correspond to different player-role outcomes.

This outcome is **not treated as independently annotated V1 ground truth**. It is derived from the verified movement labels and the rule convention encoded in the baseline implementation.

This distinction is important because the encoded rule should not be presented as the only way Ampe is played or interpreted across Ghana. The outcome measure is used only to examine whether a movement-classification error changes the benchmark-level interpretation of an exchange.

Consequently, the primary supervised targets in Ampe-DB V1 remain:

1. four-class movement configuration; and
2. binary movement relation derived from that configuration.

### 3.5 Why Ampe Presents a Distinct Recognition Problem

The Ampe interaction contains several properties that make it useful for human-activity and interaction analysis.

First, the visual differences between classes can be fine-grained. Much of the scene, including the participants, clothing, background, and general jumping motion, may remain similar while the target changes according to the leg configuration.

Second, the activity is dyadic. The complete target depends on both participants rather than on the movement of one person in isolation.

Third, the interaction is temporal. The most discriminative configuration emerges within a short movement cycle involving preparation, jumping, leg selection, and landing.

Fourth, repeated participants and partner combinations create a generalization challenge. A model may potentially exploit visual information associated with particular people or interaction contexts unless evaluation explicitly separates these factors.

These characteristics motivate the two benchmark tasks and the leakage-aware evaluation protocols formalized in Section 6.

# 4. Ampe-DB Dataset Construction

## 4.1 Collection Setup

Ampe-DB was constructed from two-person Ampe recordings involving **15 consenting participants**.

Project documentation records a **single-camera frontal recording design** in which both players were maintained within the same field of view and their full-body movements remained visible throughout the interaction. The documentation identifies **Blackmagic Pocket Cinema Camera 6K Pro** and **Sony A7 II** cameras as the recording devices used across the collection.

The recordings were conducted in semi-controlled environments intended to preserve relatively natural gameplay while maintaining sufficient visibility for subsequent 2D pose extraction. Both indoor and outdoor conditions are represented in the project documentation.

The collection was designed around lightweight monocular video rather than multi-camera motion capture, depth sensing, wearable sensors, or body-mounted markers.

Because the final provenance audit did not recover a sufficiently reliable mapping from every canonical clip to a verified recording-session identifier, Ampe-DB V1 does **not** define or release an official session-based benchmark split. Collection conditions are therefore described from the project documentation, while clip-level technical properties are reported from the final media audit.

## 4.2 Segmentation and Canonical Clips

Longer gameplay recordings were segmented into short clips corresponding to individual interaction exchanges.

Following annotation, media auditing, identity reconciliation, pose alignment, and canonicalization, Ampe-DB V1 contains **637 canonical clips**.

The final canonical video collection has a total duration of approximately **19.5 minutes**, with a mean clip duration of approximately **1.84 seconds**.

The clips contain between **27 and 94 video frames**, with a median of approximately **60 frames**.

The canonical clips are stored at a resolution of **1920 × 1080 pixels**.

Frame rate is not uniform across the final dataset. The technical audit identifies:

- **623 clips at 30 fps**;
- **12 clips at 24 fps**; and
- **2 clips at 60 fps**.

The average frame rate across the complete canonical collection is approximately **29.98 fps**.

This variability is preserved in the released media rather than artificially standardizing the source files. Baseline preprocessing therefore samples temporal positions relative to each clip's own source-video timeline.

### 4.3 Movement-Class Distribution

The 637 canonical clips are distributed across the four verified movement configurations as follows:

| **Movement configuration** | **Clips** | **Percentage** |
|---|---|---|
| left_left | 200 | 31.40% |
| left_right | 134 | 21.04% |
| right_left | 114 | 17.90% |
| right_right | 189 | 29.67% |
| **Total** | **637** | **100%** |

The distribution is not perfectly balanced, which motivates the use of macro-F1 as the primary evaluation metric and class-weighted loss during baseline training.

### 4.4 Available Modalities

Each canonical Ampe-DB record links multiple representations of the same two-person interaction.

The principal source modality is **RGB video**. The canonical video files also contain synchronized embedded audio, with the V1 media audit recording the primary audio stream as **AAC stereo at 44.1 kHz**. Audio is retained as part of the source dataset but is **not used in the Baseline V1 experiments** reported in this paper.

A derived **two-person pose modality** is additionally provided. Initial pose extraction produced frame-level body-landmark data in CSV and JSON formats together with rendered skeleton videos. These skeleton videos provide a visual representation of the extracted body landmarks and were retained primarily for visual inspection, quality control, and dataset illustration.

The raw pose outputs were subsequently processed, normalized, audited, and aligned with the canonical RGB records. The resulting normalized 2D pose files constitute the **canonical pose representation used by Baseline V1**. They preserve the two-player structure of each interaction while reducing much of the appearance and background information present in the RGB recordings.

Accordingly, the benchmark treats **RGB and pose as the two evaluated visual modalities**. The rendered skeleton videos are visualizations of the pose information rather than a separate input modality, while audio remains an available but unevaluated source modality in V1.

**Table 1. Overview of Ampe-DB V1**

| **Property** | **Ampe-DB V1** |
|---|---|
| Canonical interaction clips | 637 |
| Participants | 15 |
| Unordered participant pairs | 11 |
| Unique audiovisual SHA-256 hashes | 557 |
| Total canonical video duration | ≈19.5 min |
| Mean clip duration | ≈1.84 s |
| Clip frame range | 27–94 frames |
| Median frames per clip | 60 |
| Canonical RGB resolution | 1920 × 1080 |
| Frame rates | 623 × 30 fps; 12 × 24 fps; 2 × 60 fps |
| RGB modality | Monocular video |
| Audio | AAC stereo, 44.1 kHz |
| Pose modality | Normalized two-person 2D pose |
| Verified semantic label family | Four-class movement configuration |
| Derived relational target | Same-leg / different-leg configuration |
| Four-class distribution | 200 / 134 / 114 / 189 |
| Baseline protocols | Duplicate-safe fixed; pair-held-out; participant-held-out diagnostic |

### 4.5 Person Detection and Pose Extraction

The pose-processing pipeline was designed to produce temporally aligned skeletal representations for both participants.

Person regions were first detected using the YOLOv8n object-detection model (Jocher et al., 2023). Detected player regions were then associated across consecutive frames using spatial continuity and positional consistency before pose estimation.

Two-dimensional body landmarks were extracted using MediaPipe Pose (BlazePose) (Lugaresi et al., 2019; Bazarevsky et al., 2020). Project documentation records the following configuration:

- model_complexity = 1;
- minimum detection confidence = 0.5;
- minimum tracking confidence = 0.5; and
- temporal landmark smoothing enabled.

MediaPipe provides 33 body landmarks for each represented person. The resulting paired pose sequences contain body-joint information for the two player slots across the interaction.

The extracted keypoints were subsequently normalized using body-centred alignment based on torso reference joints, including the shoulders and hips. This normalization reduces variation associated with absolute image position, participant scale, and spatial displacement while preserving the left/right structure required by the benchmark labels.

The V1 pose audit separately verifies the structural integrity, source-frame coverage, landmark coverage, and visibility statistics of the canonical pose files.

### 4.6 Temporal Alignment and Missing Pose Observations

Baseline V1 does not independently resample only those frames in which pose estimation succeeds.

Instead, pose input positions are selected relative to the **source-video frame timeline**. Thirty-two temporal positions are sampled deterministically for baseline modelling.

When the pose representation for a player is missing at a selected source-frame position, the corresponding coordinates are zero-filled and an explicit player-presence indicator records the missing observation.

This preserves temporal alignment between the two participants and avoids implicitly compressing time when a detection is absent.

### 4.7 Canonical Manifest

The different representations of each interaction are connected through a canonical manifest.

For every record, the manifest links information including:

- canonical clip identifier;
- RGB video path;
- pose filename;
- frame count;
- frame rate and duration;
- four-class movement configuration;
- derived movement relation;
- participant identifiers;
- unordered participant-pair identifier;

- audiovisual SHA-256 hash; and
- technical and provenance quality-control fields.

The final manifest contains **637 unique canonical record identifiers**.

The canonical manifest serves as the authoritative dataset index used during benchmark construction. Labels are therefore not reconstructed from file or directory names during experimentation.

### 4.8 Exact-Content Auditing

SHA-256 hashing of the complete canonical audiovisual files identifies **557 unique content hashes among the 637 records**.

This demonstrates that some records contain byte-identical audiovisual content under different canonical record identities.

The duplicates are documented rather than silently ignored. Benchmark construction groups records at the SHA-256 level so that identical audiovisual content cannot occur across train, validation, and test partitions.

This distinction between record identity and content identity is important for reproducibility and is carried through all three Baseline V1 evaluation protocols.

### 4.9 Participant and Pair Structure

The 15 participants represented in Ampe-DB contribute interactions across **11 unordered participant pairs**.

This dyadic structure creates dependencies beyond ordinary clip-level repetition. The same participant may occur in multiple clips and may appear with different partners, while the same participant pair may contribute repeated interaction examples.

These properties directly motivate the pair-held-out and participant-held-out protocols introduced in Section 6.

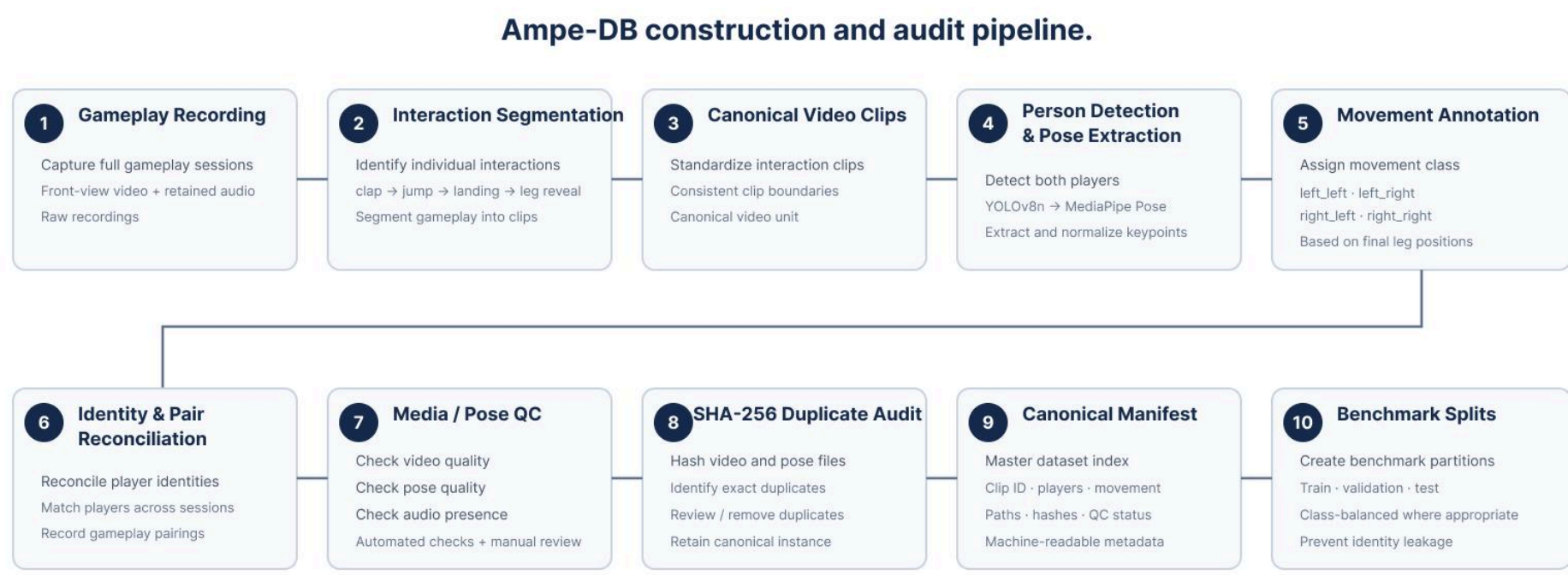

**Figure 3. Ampe-DB construction and audit pipeline.** Gameplay recording → interaction segmentation → canonical video clips → person detection and pose extraction → movement annotation → identity and pair reconciliation → media/pose quality control → SHA-256 duplicate audit → canonical manifest → benchmark splits.

## 4.10 Final V1 Dataset

The final Ampe-DB V1 collection therefore comprises **637 canonical two-person interaction clips from 15 participants**, representing **11 unordered participant pairs** and **557 unique audiovisual content hashes**.

The dataset provides RGB video, synchronized embedded audio, normalized two-person 2D pose, verified four-class movement configurations, derived binary movement relations, participant and pair metadata, and the quality-control information required to construct leakage-aware evaluation protocols.

These properties form the basis for the annotation audit and benchmark procedures described in the following sections.

# 5. Annotation, Audit, and Quality Control

## 5.1 Annotation Schema

Ampe-DB separates verified semantic annotations from labels and metadata derived for benchmark evaluation. The primary semantic ground-truth field is the verified four-class movement configuration defined in Section 3:

$$Y_4 = \{left_left, left_right, right_left, right_right\}.$$

Each class records the stored two-token movement configuration associated with a canonical interaction record. The stored token order distinguishes left_right from right_left, but the V1 audit does not independently establish a mapping from movement-token order to the canonical `player_1_id` and `player_2_id` fields. The binary movement relation is not treated as an independently annotated target. Instead, it is derived deterministically from the verified four-class configuration according to the mapping defined in Section 3.3. Participant identifiers and unordered participant-pair identifiers are stored separately as metadata to support participant-aware and pair-aware evaluation.

**Table 2. Core annotation and benchmark fields in the Ampe-DB canonical manifest**

| Field | Description | Role |
|---|---|---|
| clip_id | Unique canonical record identifier | Record identity |
| movement_class | Verified four-class movement configuration | Primary semantic target |
| movement_relation | Derived same-leg/different-leg relation | Derived binary target |
| player_1_id | Canonical pseudonymous Player 1 participant identifier | Participant metadata |
| player_2_id | Canonical pseudonymous Player 2 participant identifier | Participant metadata |
| pair_id_unordered | Participant-pair identity independent of order | Pair-level evaluation |
| video_relative_path | Location of the corresponding RGB video | Cross-modal alignment |
| pose_filename | Corresponding pose representation | Cross-modal alignment |
| frame_count | Number of source-video frames | Temporal metadata |
| sha256 | Cryptographic hash of audiovisual content | Duplicate auditing |

The canonical manifest is the authoritative benchmark index. Experimental code retrieves labels, participant metadata, and media references from this manifest rather than reconstructing them from file or directory names.

### 5.2 Canonicalization and Media Integrity

Before benchmark construction, intermediate outputs from segmentation, pose extraction, annotation, identity reconciliation, and quality review were consolidated into a single canonical record table.

The final audit required every benchmark record to satisfy three basic integrity conditions:

1. a unique canonical clip identifier;
2. a valid corresponding RGB video and pose representation; and
3. alignment between the media references, semantic labels, and participant metadata stored in the canonical manifest.

All 637 canonical records satisfied the identifier requirement, and all 637 required RGB-video and pose paths were successfully resolved.

The canonical manifest was frozen before execution of the final benchmark experiments. This ensured that subsequent training and evaluation used a single, stable mapping between records, labels, participants, and modalities.

### 5.3 Exact-Content Duplicate Audit

Exact audiovisual duplicates were identified using SHA-256 hashing of the complete canonical video files. Among the 637 canonical records, the audit identified 557 distinct SHA-256 values, showing that some canonical record identifiers correspond to byte-identical audiovisual content.

These records were retained because canonical record identity and audiovisual-content identity represent different dataset properties. Duplicate information was instead incorporated directly into benchmark construction.

For every evaluation protocol, all records sharing the same SHA-256 hash were constrained to the same train, validation, or test partition. Consequently, an exact audiovisual duplicate of a test example cannot appear in the corresponding model-development data.

The duplicate-group information is retained as part of the benchmark artifacts so that the leakage controls can be reproduced in future experiments.

### 5.4 Player-Slot and Label Consistency

The four-class movement target preserves a stored two-token ordering. In particular, left_right and right_left contain the same two token values in different stored orders and are therefore retained as distinct classes.

The V1 audit does not independently establish that this movement-token ordering corresponds to the canonical `player_1_id` and `player_2_id` ordering. Pose records preserve two within-file player channels, and these channels are handled consistently when model inputs are assembled. However, the temporary pose-tracking identifiers are distinct from the persistent participant identities stored in the canonical manifest and should not be interpreted as establishing the semantic ordering of the movement tokens.

A separate consistency check verifies the relationship between the four-class movement label and the derived binary target. For every canonical record:

$$\{left_left, right_right\} \rightarrow same_leg_configuration$$

while:

$$\{left_right, right_left\} \rightarrow different_leg_configuration.$$

The binary target is therefore derived from the verified four-class movement configuration rather than maintained as an independent semantic annotation.

That distinction also agrees with the Pose Format, which explicitly says pose-file player IDs should be kept separate from canonical participant identity and movement-token semantic ordering.

## 5.5 Participant-Class Coverage Audit

Participant-held-out evaluation requires sufficient target-class coverage for the participant used as the test subject. A complete participant-by-class audit was therefore performed across all 15 participants before the diagnostic protocol was finalized.

For four-class evaluation, a participant is eligible only if their associated recordings contain all four movement configurations. Six participants satisfy this requirement:

{P01,P02,P03,P04,P06,P07}

For binary movement-relation evaluation, the participant must contain both same-leg and different-leg examples. Eight participants satisfy this requirement:

{P01,P02,P03,P04,P06,P07,P11,P12}

**Table 3. Participant eligibility for held-out diagnostic evaluation**

| Participant | Four-class eligible | Binary eligible | Reason when ineligible |
|---|---|---|---|
| P01 | Yes | Yes | — |
| P02 | Yes | Yes | — |
| P03 | Yes | Yes | — |
| P04 | Yes | Yes | — |
| P05 | No | No | Missing both cross-leg configurations; no different-leg examples |
| P06 | Yes | Yes | — |
| P07 | Yes | Yes | — |
| P08 | No | No | Missing both cross-leg configurations; no different-leg examples |
| P09 | No | No | Missing both cross-leg configurations; no different-leg examples |
| P10 | No | No | Missing both cross-leg configurations; no different-leg examples |
| P11 | No | Yes | Missing left_left for four-class evaluation |
| P12 | No | Yes | Missing left_left for four-class evaluation |
| P13 | No | No | Missing left_left and right_right; no same-leg examples |
| P14 | No | No | Only right_right represented |
| P15 | No | No | Only right_right represented |

Participants that do not satisfy these requirements remain part of Ampe-DB. They are excluded only as held-out diagnostic targets for tasks in which their class coverage is incomplete and may still contribute to training or validation where the identity-separation constraints permit.

This eligibility rule prevents a participant-held-out fold from being interpreted as a complete classification test when one or more target categories are structurally absent from that participant's recordings.

### 5.6 Split-Level Leakage Validation

After the benchmark partitions were constructed, each protocol underwent a final leakage audit.

For the duplicate-safe fixed benchmark, every SHA-256 content group was verified to occur in only one of the train, validation, or test partitions.

For each pair-held-out fold, two conditions were checked:

1. no exact-content hash crossed partition boundaries; and
2. every participant pair assigned to the test set was absent from both training and validation.

For participant-held-out evaluation, the held-out participant was required to be absent from both training and validation, while every test clip in that fold was required to contain the held-out participant.

All locked Baseline V1 folds passed these leakage checks.

### 5.7 Benchmark Freeze and Reproducibility Audit

Following execution of the duplicate-safe fixed, pair-held-out, and participant-held-out experiments, the complete Baseline V1 package underwent a final reproducibility audit.

The audit covered dataset cardinality, cache completeness, experiment counts, prediction-file sizes, fusion-selection outputs, training histories, duplicate separation, participant separation, participant eligibility, confusion-matrix normalization, and the presence of required reporting artifacts.

A total of 35 final integrity checks were executed, with zero failures.

The resulting benchmark was subsequently frozen as **Ampe-DB Baseline V1**. The freeze preserves the originally executed test results and prevents model-selection procedures, test partitions, or reported predictions from being changed after test performance had been observed.

This distinction is particularly important for the multimodal fusion analysis reported later in the paper. Post-hoc analysis identified cases in which validation-based fusion-weight selection was unstable under participant shift, but these observations were used only to interpret the frozen benchmark results and did not alter them.

# 6. Benchmark Tasks and Evaluation Protocols

## 6.1 Evaluation Design

Ampe-DB is designed to evaluate not only whether a model can recognize movement configurations, but also whether the learned representation transfers beyond the people and interaction pairings observed during model development.

This distinction is important because multiple clips may contain the same participant or participant pair. A conventional clip-level split can therefore place related interaction contexts across development and test data even when exact duplicate clips are excluded.

To examine different forms of generalization, Ampe-DB defines three complementary evaluation protocols:

1. a duplicate-safe fixed benchmark;
2. a pair-held-out benchmark; and
3. a participant-held-out diagnostic evaluation.

All three protocols operate on the same canonical Ampe-DB records and use the same task definitions. Exact audiovisual duplicates identified by SHA-256 are prevented from crossing train, validation, and test partitions in every protocol, as described in Section 5.

**Table 4. Ampe-DB benchmark evaluation protocols**

| **Protocol** | **Main question** | **Test-set separation** | **Aggregation unit** | **Role in benchmark** |
|---|---|---|---|---|
| Duplicate-safe fixed | Can the model recognize Ampe movements under the predefined benchmark distribution? | Exact duplicate content cannot cross partitions | Training seed | Reference benchmark |
| Pair-held-out | Does the model generalize to participant pairings not observed during development? | Test participant pairs are absent from training and validation | Held-out pair fold | Primary generalization benchmark |
| Participant-held-out | Does the model generalize when a participant has not appeared during development? | Held-out participant is absent from training and validation | Held-out participant | Identity-generalization diagnostic |

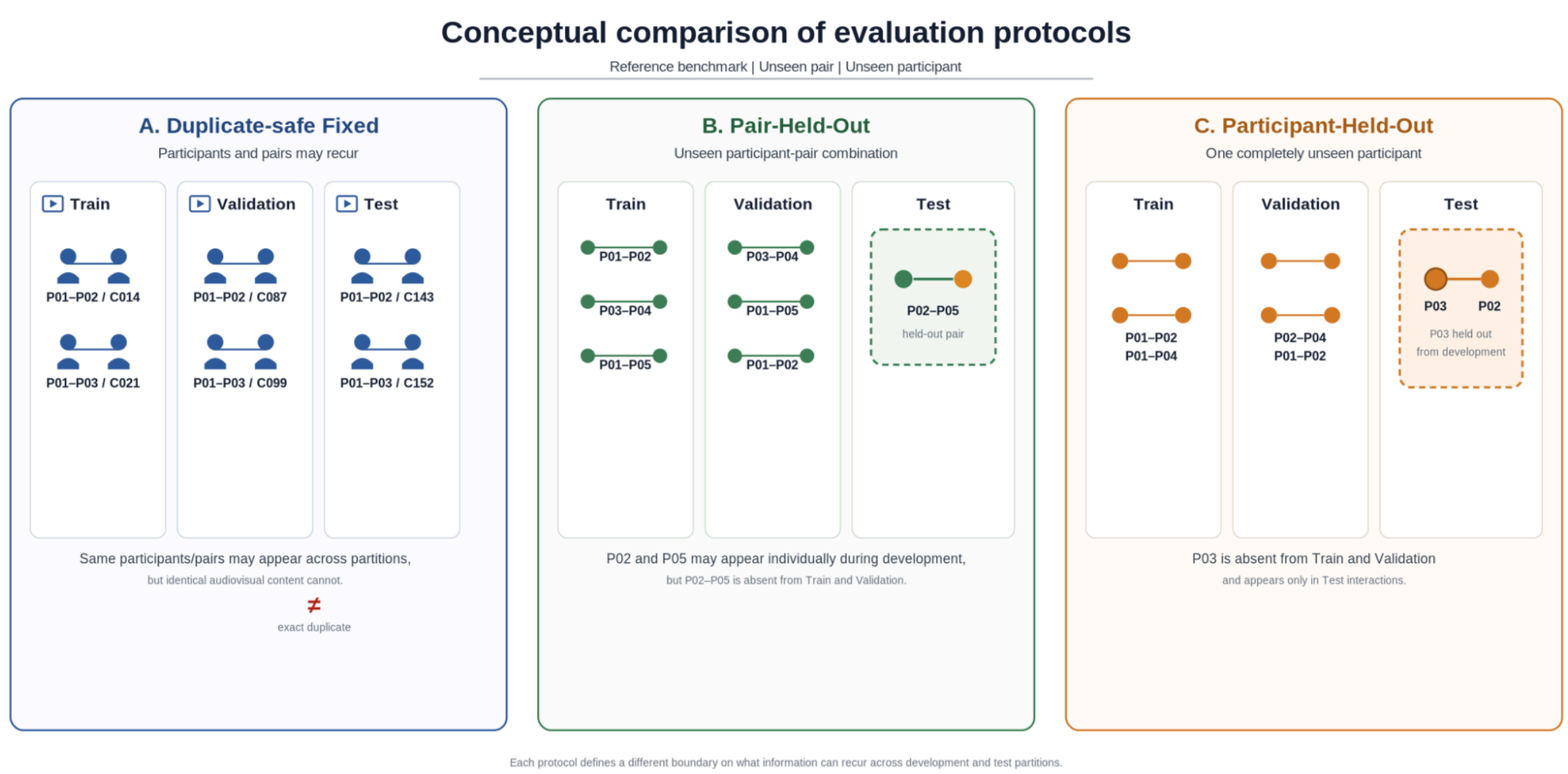

**Figure 4. Ampe-DB evaluation protocols.** Schematic comparison of the three evaluation settings. In the duplicate-safe fixed benchmark, participants and participant pairs may occur across development and test partitions, but identical audiovisual content cannot cross partitions. In pair-held-out evaluation, participant pairs assigned to the test fold are absent from training and validation. In participant-held-out evaluation, the target participant is excluded entirely from model development.

## 6.2 Benchmark Tasks

Ampe-DB evaluates the two supervised classification tasks defined in Section 3.

### Four-Class Movement Recognition

Given an Ampe interaction clip x, the model predicts

$\hat{y}_4 = f(x)$

where

$Y_4 = \{\text{left_left}, \text{left_right}, \text{right_left}, \text{right_right}\}$

Because stored token order is part of the four-class label, left_right and right_left are treated as distinct classes. This distinction does not imply a verified correspondence between token position and the canonical Player 1 / Player 2 participant ordering.

**Binary Movement-Relation Classification**

The second task predicts

$\hat{y}_2 = g(x)$

where

$Y_2 = \{\text{same_leg_configuration}, \text{different_leg_configuration}\}$

The binary task removes the specific left/right ordering while retaining the relationship between the two participants' movements.

The same three evaluation protocols are applied to both tasks, subject to the participant-class coverage requirements of the participant-held-out diagnostic.

## 6.3 Duplicate-Safe Fixed Benchmark

The duplicate-safe fixed benchmark provides a reference condition for comparing modalities and model configurations under one predefined train, validation, and test partition.

Let h(x) denote the SHA-256 hash of clip x, and let s(x) denote its assigned partition. For every audiovisual-content hash h, $|\{s(x) : h(x) = h\}| = 1$

Thus, all canonical records sharing identical audiovisual content are assigned to the same partition.

The fixed benchmark does not require participant identities or participant pairs to be mutually exclusive between development and test data. It therefore measures recognition under duplicate-safe but comparatively less restrictive distributional conditions.

Models are trained using the training partition. Checkpoint selection and fusion decisions are made using validation data, and the test partition is reserved for final evaluation.

Because the partition itself is fixed, variability for this protocol is estimated across the predefined training seeds.

### 6.4 Pair-Held-Out Benchmark

The pair-held-out benchmark is the principal generalization evaluation in Ampe-DB.

Because Ampe is a dyadic activity, repeated interactions involving the same two participants may share characteristics beyond the target movement configuration. The pair-held-out protocol therefore evaluates whether a representation transfers to participant pairings that were not observed during model development.

The canonical dataset is evaluated using four outer folds. For fold k, let $\Pi_{test}^{(k)}$ denote the set of unordered participant pairs assigned to the test partition. The protocol requires:

$$\Pi_{test}^{(k)} \cap \Pi_{train}^{(k)} = \varnothing$$

and

$$\Pi_{test}^{(k)} \cap \Pi_{validation}^{(k)} = \varnothing.$$

Exact duplicate separation is additionally maintained within each fold.

Importantly, the protocol holds out **pair identities**, not necessarily every individual participant belonging to those pairs. A participant may therefore appear in development data with one partner and in the test set with another. The protocol consequently measures transfer to unseen participant combinations rather than complete transfer to unseen individuals.

Each of the four outer folds is evaluated independently.

For a metric m, results are first averaged across the three training seeds within fold k:

$$\bar{m}_k = (1/S) \sum_{s=1}^{S} m_{(k,s)},$$

where

$$S = 3.$$

The overall pair-held-out score is then the unweighted mean across the four outer folds:

$\bar{m}_{pair} = (1/4) \sum_{k=1}^{4} \bar{m}_k$.

Treating folds equally prevents folds containing more test clips from automatically contributing more weight to the reported benchmark score.

### 6.5 Participant-Held-Out Diagnostic Evaluation

The participant-held-out protocol examines transfer when one participant in the test interactions has been completely excluded from model development.

For an eligible held-out participant p,

$p \notin P_{train}$

and

$p \notin P_{validation}$,

while every test clip contains that participant:

$p \in P(x)$, for all $x \in D_{test}$,

where P(x) denotes the set of participants appearing in clip x.

The protocol therefore asks whether a model can recognize an Ampe movement when one of the people performing it has never appeared during training or validation.

Participant-held-out evaluation is treated as a diagnostic rather than the principal benchmark because complete target-class coverage is not available for all 15 participants.

As established by the participant-class coverage audit in Section 5, **six participants** are eligible as held-out targets for four-class recognition and **eight participants** are eligible for binary movement-relation classification.

Participants lacking complete target-class coverage are not used as held-out test targets for the corresponding task. Their recordings remain part of Ampe-DB and may contribute to development data in folds where they do not violate the held-out participant constraint.

Exact audiovisual duplicate separation is maintained within every participant-held-out fold.

### 6.6 Aggregation of Participant-Held-Out Results

Participant-held-out folds overlap in their development data and differ in the number of available test clips. Predictions from all held-out participants are therefore not pooled into a single combined test set.

Instead, performance is first averaged across the three predefined training seeds for each eligible held-out participant p:

$\bar{m}_p = (1/S) \Sigma_{(s=1)}^S m_{(p,s)}$,

where

$S = 3$.

The overall participant-held-out diagnostic score is then the unweighted mean across eligible participants:

$\bar{m}_{participant} = (1/N) \Sigma_{(p=1)}^N \bar{m}_p$, where

$N = 6$

for four-class recognition and

$N = 8$

for binary movement-relation classification.

This aggregation preserves the held-out participant as the unit of analysis and prevents participants with more recorded clips from receiving disproportionate weight in the final diagnostic score.

Variability is reported across the participant-level seed-averaged scores.

### 6.7 Training Seeds and Repeated Evaluation

All trainable baseline models are evaluated using three predefined random seeds:

{17, 29, 43}.

The seed controls stochastic components of model initialization and training while leaving the benchmark partitions unchanged.

The same seeds are used across modalities, tasks, and evaluation protocols.

Results are summarized differently according to the evaluation unit:

- **duplicate-safe fixed:** variation across the three training seeds;
- **pair-held-out:** seeds are averaged within each outer fold, with variation reported across the four fold-level scores;
- **participant-held-out:** seeds are averaged within each held-out participant, with variation reported across eligible participant-level scores.

This preserves the appropriate unit of analysis for each protocol.

### 6.8 Evaluation Metrics

The primary evaluation metric is macro-averaged F1 score.

For C target classes:

$$F1_macro = (1/C)\ \Sigma_(c=1)\hat{}C\ F1_c.$$

Macro-F1 gives equal weight to each class regardless of its frequency. This is important for Ampe-DB because the four-class distribution is not perfectly balanced and participant-level class coverage varies.

Accuracy is reported as a complementary metric:

Accuracy = number of correctly classified clips / number of evaluated clips.

Per-class precision, recall, and F1 are retained for diagnostic analysis.

Confusion matrices are row-normalized so that each row represents the distribution of predictions for a given true class. For held-out protocols, normalized confusion matrices are first constructed at the corresponding evaluation unit and then averaged using the same fold-aware or participant-aware aggregation procedure as the primary metrics.

Where Baseline V1 reports rule-derived outcome accuracy, the outcome is computed deterministically from the predicted movement state and the encoded benchmark rule. It is treated as an auxiliary interpretation measure rather than an independently annotated recognition target.

### 6.9 Model Selection and Test Isolation

All model-selection decisions are restricted to training and validation data.

Test partitions are not used to select model checkpoints, fusion weights, preprocessing choices, or hyperparameters during the locked Baseline V1 benchmark.

For multimodal late fusion, candidate RGB-pose fusion weights are evaluated using the corresponding validation partition. The selected weight is then applied unchanged to test predictions.

After completion of the benchmark, diagnostic analysis identified instability in this validation-based fusion procedure for some participant-held-out binary evaluations. These observations were used only to interpret the frozen results and were not used to revise the selected weights or replace the original test predictions.

This separation preserves the role of the test partitions as held-out evaluation data.

### 6.10 Interpretation of the Three Protocols

The three evaluation protocols measure different forms of generalization and should not be interpreted as interchangeable estimates of the same quantity.

The duplicate-safe fixed benchmark measures recognition under the predefined benchmark distribution while controlling exact-content leakage.

The pair-held-out benchmark evaluates transfer to participant combinations that were absent from model development.

The participant-held-out diagnostic evaluates transfer when a specific participant has not appeared in training or validation.

Performance differences across these settings therefore provide information about how sensitive a representation is to changes in participant and interaction structure. A reduction in performance under held-out evaluation can indicate that information that was predictive within the fixed benchmark transfers less reliably when familiar participant pairings or identities are removed.

# 7. Baseline Models and Experimental Setup

## 7.1 Baseline Design

The Ampe-DB baseline suite compares three complementary representations of the interaction:

1. RGB appearance and motion, represented using pretrained video features;
2. body-pose dynamics, represented using temporally ordered two-person pose sequences; and
3. multimodal RGB-pose information, represented using probability-level late fusion.

A majority-class predictor is additionally included as a non-learned reference baseline.

The purpose of Baseline V1 is not to establish state-of-the-art performance on Ampe-DB. Instead, the models provide reproducible reference points for examining the relative behaviour of RGB, pose, and their combination under the evaluation protocols defined in Section 6.

The same model definitions and optimization settings are used for both benchmark tasks and across the three evaluation protocols.

## 7.2 Majority-Class Baseline

For each training partition, the majority baseline identifies the most frequent target class:

$c^* = \arg\max_c N_c$,

where $N_c$ is the number of training examples belonging to class c.

Every test example is subsequently assigned to c*.

The majority class is determined independently for each task and benchmark partition using training data only. This provides a non-learned reference against which the RGB, pose, and multimodal models can be compared.

### 7.3 RGB Representation

RGB information is represented using an **R(2+1)D-18** video network (Tran et al., 2018) pretrained on **Kinetics-400** (Carreira and Zisserman, 2017).

The pretrained backbone is used as a frozen feature extractor rather than fine-tuned end-to-end on Ampe-DB. This provides a controlled baseline for examining how a generic pretrained video representation transfers to the Ampe recognition tasks without adapting the backbone to the dataset.

For each clip, **16 frames** are sampled deterministically and approximately uniformly over the complete source-video timeline.

If a source clip contains T frames, the temporal sample locations are distributed over:

$[0, T - 1]$.

The sampled frames are processed using the official preprocessing associated with R2Plus1D_18_Weights.DEFAULT in torchvision, including the resize, crop, value scaling, and normalization expected by the pretrained network.

Horizontal flipping is disabled because left and right carry semantic meaning in the Ampe movement labels. A horizontal flip could alter the apparent movement class without a corresponding transformation of the target label.

The final classification layer of R(2+1)D-18 is replaced with an identity mapping, producing a **512-dimensional feature vector** for each clip.

Because the video backbone is frozen, RGB features are extracted once for each unique audiovisual content hash and cached for reuse across evaluation folds, tasks, and training seeds.

A lightweight multilayer perceptron is trained on the resulting feature vector. The classifier consists of:

- a fully connected layer from 512 to 128 dimensions;
- ReLU activation;
- dropout with probability 0.30; and
- a task-specific output layer.

The RGB pipeline can therefore be summarized as:

$x_{RGB} \rightarrow$ frozen R(2+1)D-18 $\rightarrow z_{RGB} \rightarrow$ MLP $\rightarrow \hat{y}$,

where

$z_{RGB} \in \mathbb{R}^{512}$.

The output dimension is four for movement-configuration recognition and two for binary movement-relation classification.

### 7.4 Pose Representation

Pose inputs are represented as temporally ordered sequences containing information from both participants.

For each clip, **32 temporal positions** are sampled deterministically over the original source-video timeline. At every sampled position, the representation contains pose information for two player slots.

Each participant contributes **33 body landmarks**, with every landmark represented by:

$(x, y, v)$,

where x and y are the stored normalized coordinates and v is landmark visibility.

For two participants, this produces:

$2 \times 33 \times 3 = 198$

pose values at each temporal position.

Two additional binary indicators specify whether a complete pose observation is available for each player. The final input dimension per temporal step is therefore:

$198 + 2 = 200$.

Each pose input consequently has the form:

$X_pose \in \mathbb{R}^{(32 \times 200)}$.

When a complete player-frame pose observation is unavailable, the corresponding coordinate and visibility values are zero-filled and the associated player-presence indicator records the missing observation, following the temporal-alignment procedure described in Section 4.

### 7.5 Bidirectional GRU Pose Classifier

The pose sequences are classified using a **two-layer bidirectional gated recurrent unit (BiGRU)** network, using gated recurrent units as introduced by Cho et al. (2014).

The model receives a 200-dimensional pose vector at each of the 32 sampled temporal positions. Each GRU direction has a hidden size of **128**.

The sequence is processed in both forward and backward temporal directions. The final hidden states are concatenated:

$z_pose = [h_forward ; h_backward]$.

The resulting pose representation therefore has dimension:

256.

A dropout layer with probability **0.30** is applied before the final linear classification layer.

The pose model can be summarized as:

$X_pose \rightarrow$ two-layer BiGRU $\rightarrow z_pose \rightarrow$ Dropout $\rightarrow$ Linear $\rightarrow \hat{y}$.

The architecture is intentionally compact. Its purpose is to establish a reproducible reference for the usefulness of explicit two-person body dynamics rather than to exhaust the range of possible skeleton-based models.

### 7.6 RGB–Pose Late Fusion

The multimodal baseline combines the independently trained RGB and pose classifiers at the probability level.

Let $p_{RGB}(y \mid x)$

and

$p_{pose}(y \mid x)$

denote the class-probability distributions produced by the two models.

The fused prediction distribution is:

$p_{fusion} = \alpha p_{RGB} + (1 - \alpha) p_{pose}$.

The fusion coefficient is selected from the predefined grid:

$\alpha \in \{0, 0.25, 0.50, 0.75, 1.00\}$.

For every benchmark instance and training seed, all candidate values are evaluated using the corresponding validation partition. The value producing the highest validation macro-F1 is selected and subsequently applied unchanged to the test predictions.

Under this formulation:

- $\alpha = 0$ corresponds to pose-only prediction;
- $\alpha = 1$ corresponds to RGB-only prediction; and
- intermediate values combine the two probability distributions.

If multiple fusion coefficients obtain the same highest validation macro-F1, the implementation retains the first coefficient encountered on the ordered grid. Because the grid is evaluated from 0 to 1, an exact tie favours the more pose-weighted candidate.

No test-set information is used in fusion-weight selection.

### 7.7 Loss Function and Class Weighting

Both trainable classifiers use class-weighted cross-entropy loss.

For class c, the training weight is:

$w_c = N / (C N_c)$,

where:

- N is the total number of training examples;
- C is the number of target classes; and
- $N_c$ is the number of training examples belonging to class c.

This assigns larger loss weights to classes occurring less frequently in the corresponding training partition.

The loss is:

$\mathcal{L} = -\sum_{c=1}^{C} w_c y_c \log(p_c)$.

Class weights are recomputed independently for every benchmark training partition rather than estimated once from the complete dataset.

### 7.8 Optimization and Checkpoint Selection

The RGB classifier and pose BiGRU are optimized using **AdamW** (Loshchilov and Hutter, 2019) with:

learning rate = $10^{-3}$

and

weight decay = $10^{-4}$.

Training uses a mini-batch size of **32** and runs for a maximum of **100 epochs**.

Early stopping is applied with a patience of **12 epochs**.

Checkpoint selection is based primarily on validation macro-F1. A checkpoint replaces the current best model whenever it achieves a higher validation macro-F1.

If two epochs obtain the same validation macro-F1, the checkpoint with the lower validation loss is retained.

Only the selected validation checkpoint is used to generate test predictions.

### 7.9 Reproducibility

The three predefined training seeds specified in Section 6 are:

{17, 29, 43}.

For each run, random-number generation is initialized for Python, NumPy, and PyTorch (Paszke et al., 2019). CUDA random states are also seeded when CUDA is available.

The implementation configures deterministic cuDNN behaviour by enabling deterministic operations and disabling benchmarking where applicable.

DataLoader shuffling is controlled using a seed-specific PyTorch generator so that a training run can be reconstructed from its benchmark partition and recorded seed.

RGB feature extraction is deterministic and cached once per unique audiovisual-content hash. The cached features are subsequently reused across tasks, folds, and seeds.

Pose preprocessing is likewise cached following deterministic temporal sampling.

**Table 5. Baseline model and training configuration**

| Component | Configuration |
|---|---|
| Majority baseline | Most frequent training class |
| RGB backbone | R(2+1)D-18 |
| RGB pretraining | Kinetics-400 |
| RGB backbone training | Frozen |
| RGB frames per clip | 16 |
| RGB temporal sampling | Deterministic approximately uniform |
| RGB preprocessing | Official R2Plus1D_18_Weights.DEFAULT transform |
| RGB feature dimension | 512 |
| RGB classifier | Linear(512,128) → ReLU → Dropout → Linear |
| RGB classifier dropout | 0.30 |
| Pose temporal samples | 32 |
| Players represented | 2 |
| Pose landmarks per player | 33 |
| Landmark channels | (x), (y), visibility |
| Player-presence indicators | 2 |
| Pose input dimension per step | 200 |
| Pose model | Two-layer bidirectional GRU |
| GRU hidden size | 128 per direction |
| Pose representation dimension | 256 |
| Pose classifier dropout | 0.30 |
| Fusion | Validation-selected probability-level late fusion |
| Fusion (α) grid | 0.00, 0.25, 0.50, 0.75, 1.00 |
| Loss | Class-weighted cross-entropy |
| Optimizer | AdamW |
| Learning rate | $1 \times 10^{-3}$ |
| Weight decay | $1 \times 10^{-4}$ |
| Training batch size | 32 |
| Maximum epochs | 100 |
| Early-stopping patience | 12 epochs |
| Checkpoint criterion | Validation macro-F1; validation loss breaks ties |
| Training seeds | 17, 29, 43 |
| Horizontal flipping | Disabled |

### 7.10 Experimental Scope

The Baseline V1 suite intentionally limits architectural complexity.

The RGB backbone is frozen rather than fine-tuned end-to-end. The pose baseline uses a compact recurrent architecture rather than a specialized graph or transformer model. Multimodal integration is performed through scalar probability-level late fusion rather than a jointly learned fusion network.

These choices provide an interpretable starting point for Ampe-DB and allow the benchmark to compare generic pretrained visual features, explicit pose dynamics, and their simple combination under the same evaluation framework.

The reported results should therefore be interpreted as **reference baselines**, not as estimates of the maximum achievable performance on Ampe-DB.

# 8. Results

### 8.1 Overall Benchmark Performance

Table 6 summarizes macro-F1 across the three evaluation protocols for the majority, RGB, pose, and RGB-pose fusion baselines.

The results show a clear distinction between performance on the duplicate-safe fixed benchmark and performance when models are evaluated on unseen participant pairings or unseen participants.

**Table 6. Macro-F1 across Ampe-DB benchmark protocols**

| Task | Protocol | Majority | RGB | Pose | RGB + Pose |
|---|---|---|---|---|---|
| Four-class movement | Duplicate-safe fixed | 0.125 ± 0.000 | 0.955 ± 0.008 | 0.957 ± 0.010 | 0.976 ± 0.021 |
| Four-class movement | Pair-held-out | 0.114 ± 0.012 | 0.418 ± 0.114 | 0.811 ± 0.120 | 0.824 ± 0.108 |
| Four-class movement | Participant-held-out | 0.078 ± 0.026 | 0.384 ± 0.117 | 0.839 ± 0.108 | 0.833 ± 0.102 |
| Binary movement relation | Duplicate-safe fixed | 0.389 ± 0.000 | 0.932 ± 0.041 | 0.981 ± 0.007 | 0.989 ± 0.011 |
| Binary movement relation | Pair-held-out | 0.380 ± 0.032 | 0.625 ± 0.145 | 0.633 ± 0.156 | 0.630 ± 0.153 |
| Binary movement relation | Participant-held-out | 0.345 ± 0.024 | 0.472 ± 0.202 | 0.675 ± 0.192 | 0.575 ± 0.152 |

Values are reported as mean ± standard deviation. For the duplicate-safe fixed benchmark, variation is measured across the three training seeds. For pair-held-out evaluation, seeds are first averaged within each of the four outer folds and variability is reported across fold-level scores. For participant-held-out evaluation, seeds are first averaged within each eligible participant and variability is reported across participant-level scores.

Bold values indicate the highest mean within each task and protocol and should be interpreted descriptively rather than as tests of statistical significance.

All learned models substantially exceed the majority baseline on the duplicate-safe fixed benchmark. For four-class recognition, RGB and pose obtain similar macro-F1 values of 0.955 and 0.957, while fusion reaches 0.976. For binary movement-relation classification, RGB obtains 0.932, pose 0.981, and fusion 0.989.

The held-out protocols produce a substantially different pattern. Four-class RGB performance decreases sharply when participant-pair or participant identity structure is removed from model development, whereas the pose representation retains considerably more of its fixed-benchmark performance.

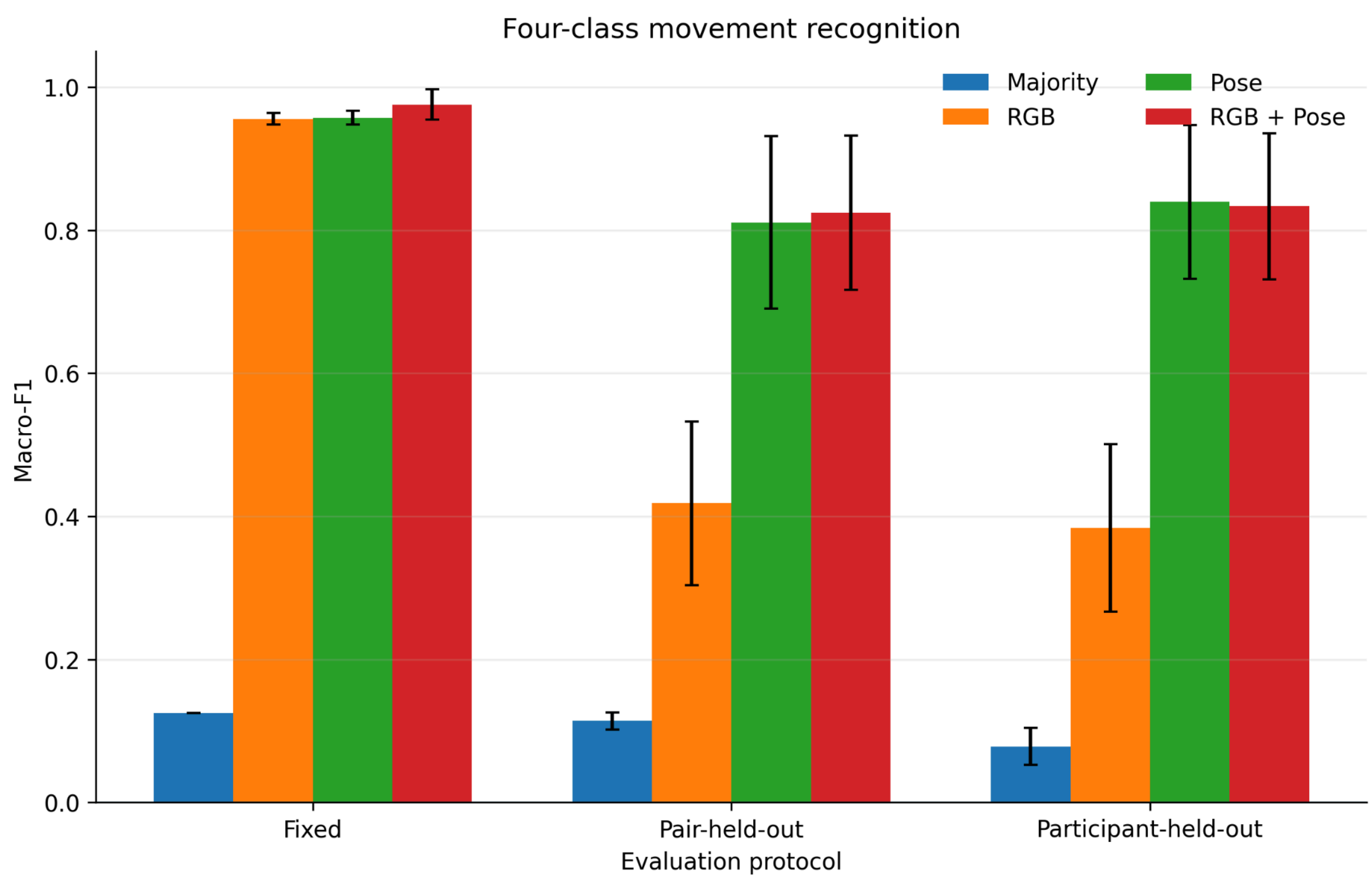

**Figure 5. Four-class movement-recognition performance across evaluation protocols.** Mean macro-F1 for the majority, RGB, pose, and RGB + pose baselines under duplicate-safe fixed, pair-held-out, and participant-held-out evaluation. Error bars represent variability according to the evaluation unit defined for each protocol.

## 8.2 Generalization to Unseen Participant Pairs

The pair-held-out benchmark produces the clearest separation between RGB and pose for four-class movement recognition.

RGB macro-F1 decreases from 0.955 on the duplicate-safe fixed benchmark to 0.418 under pair-held-out evaluation, an absolute reduction of 0.537. Accuracy decreases from 0.958 to 0.454.

Pose is considerably more stable. Four-class pose macro-F1 decreases from 0.957 to 0.811, an absolute reduction of 0.146, while accuracy decreases from 0.958 to 0.828.

RGB-pose fusion produces the highest mean four-class pair-held-out result, with macro-F1 of 0.824 and accuracy of 0.837. The improvement over pose alone is modest, approximately 0.014 macro-F1.

The four-class results therefore reveal a substantial difference between the strength of the frozen RGB representation under the fixed benchmark and its performance on participant pairings excluded from model development.

For binary movement-relation classification, the three learned representations are much closer under pair-held-out evaluation. RGB obtains 0.625 macro-F1, pose 0.633, and fusion 0.630. Their corresponding accuracies are 0.659, 0.680, and 0.678.

Despite the smaller separation between modalities, all three show substantial reductions relative to their fixed-benchmark performance. The macro-F1 reductions are 0.307 for RGB, 0.348 for pose, and 0.358 for fusion.

Thus, reducing the target space from four movement configurations to two relational classes does not eliminate the generalization difficulty associated with unseen participant pairings.

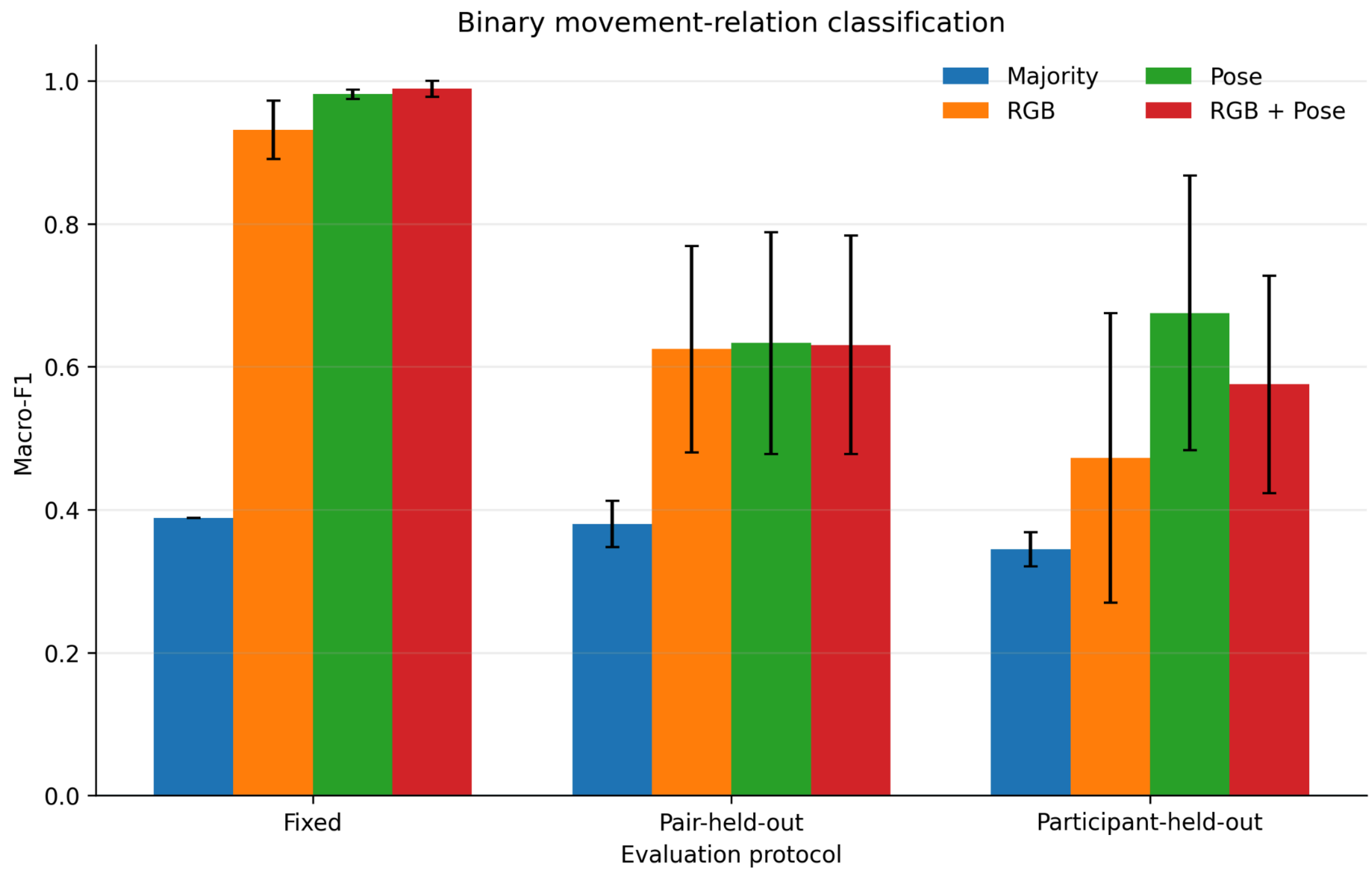

**Figure 6. Binary movement-relation performance across evaluation protocols.** Mean macro-F1 for the majority, RGB, pose, and RGB + pose baselines under the three evaluation settings. Performance decreases from the duplicate-safe fixed benchmark under both pair-held-out and participant-held-out evaluation.

### 8.3 Generalization to Unseen Participants

Participant-held-out evaluation examines performance when one participant in each test interaction has been completely absent from model development.

For four-class movement recognition, RGB obtains a mean macro-F1 of 0.384, whereas pose reaches 0.839. Their corresponding accuracies are 0.455 and 0.857.

Relative to the duplicate-safe fixed benchmark, the four-class RGB macro-F1 decreases by 0.572, while pose decreases by only 0.117.

RGB-pose fusion obtains 0.833 macro-F1 and 0.850 accuracy. It therefore closely matches, but does not exceed, pose alone. The mean difference between fusion and pose is approximately −0.006 macro-F1.

For binary movement-relation classification, pose again produces the highest mean participant-held-out result at 0.675 macro-F1 and 0.704 accuracy. RGB obtains 0.472 macro-F1, while fusion obtains 0.575.

The binary participant-held-out results differ from the four-class diagnostic in an important respect. Whereas four-class fusion remains close to pose, binary fusion performs approximately 0.100 macro-F1 below pose.

**Table 7. Change in macro-F1 relative to the duplicate-safe fixed benchmark**

| Task | Modality | Pair-held-out gap | Participant-held-out gap |
|---|---|---|---|
| Four-class | RGB | −0.537 | −0.572 |
| Four-class | Pose | −0.146 | −0.117 |
| Four-class | RGB + Pose | −0.151 | −0.142 |
| Binary relation | RGB | −0.307 | −0.459 |
| Binary relation | Pose | −0.348 | −0.306 |
| Binary relation | RGB + Pose | −0.358 | −0.414 |

Negative values indicate lower macro-F1 than on the duplicate-safe fixed benchmark.

The generalization-gap comparison is particularly striking for four-class recognition. RGB and pose are nearly indistinguishable on the fixed benchmark, differing by only 0.002 macro-F1, but their participant-held-out means differ by approximately 0.456.

## 8.4 Class-Level Error Patterns

Aggregate macro-F1 does not show which movement configurations account for the held-out performance differences. We therefore examine row-normalized confusion matrices for the pose model, which provides the strongest and most stable four-class held-out performance.

On the duplicate-safe fixed benchmark, four-class pose recall is approximately 0.92 for left_left, 0.93 for left_right, and 1.00 for both right_left and right_right.

Under pair-held-out evaluation, recall decreases most visibly for the cross-leg configurations. Recall is approximately 0.78 for left_right and 0.70 for right_left.

A similar pattern is observed under participant-held-out evaluation, where recall is approximately 0.73 for left_right and 0.79 for right_left.

The errors are not distributed uniformly across the four classes. left_right is frequently confused with left_left, while errors for right_left are distributed toward the same-leg classes, including left_left and right_right.

These results show that the cross-leg configurations become substantially more difficult under held-out evaluation than their fixed-benchmark performance suggests.

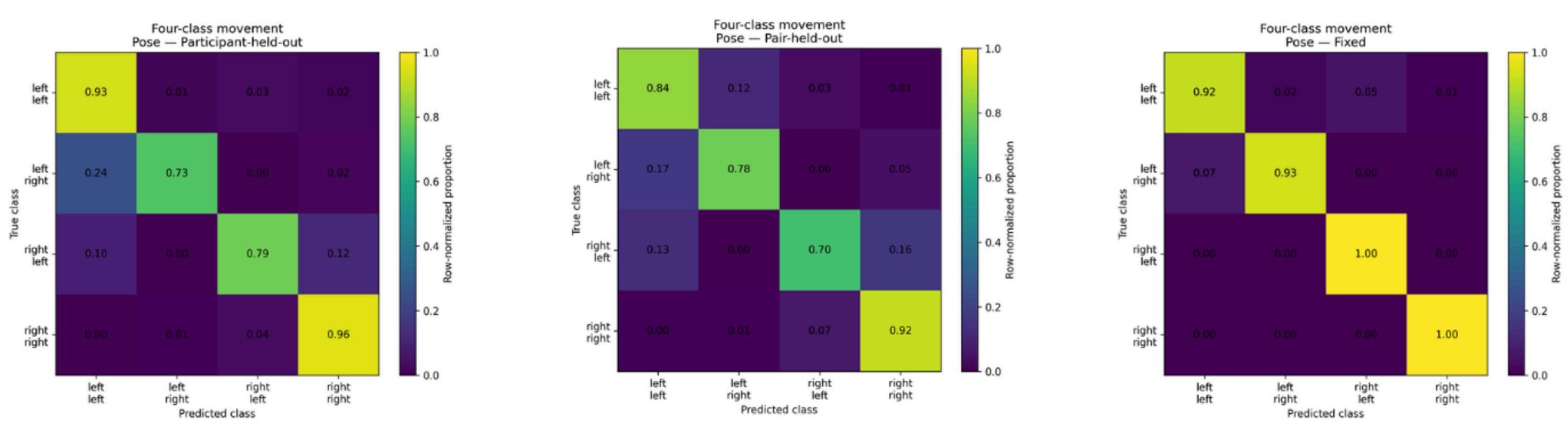

**Figure 7. Pose confusion matrices for four-class movement recognition.** Row-normalized confusion matrices for (a) duplicate-safe fixed, (b) pair-held-out, and (c) participant-held-out evaluation. The held-out protocols show increased confusion for the left_right and right_left configurations.

## 8.5 Binary Relation Errors

A related pattern appears in the binary pose results.

On the duplicate-safe fixed benchmark, recall is approximately 0.98 for both different_leg_configuration and same_leg_configuration.

Under pair-held-out evaluation, recall for different_leg_configuration decreases to approximately 0.48, whereas recall for same_leg_configuration remains substantially higher at approximately 0.84.

Participant-held-out evaluation shows a similar asymmetry. Recall is approximately 0.52 for different_leg_configuration and 0.89 for same_leg_configuration.

The binary held-out errors are therefore concentrated disproportionately in the different-leg relation.

This pattern is consistent with the four-class confusion matrices, where the two cross-leg configurations, left_right and right_left, show larger reductions under held-out evaluation.

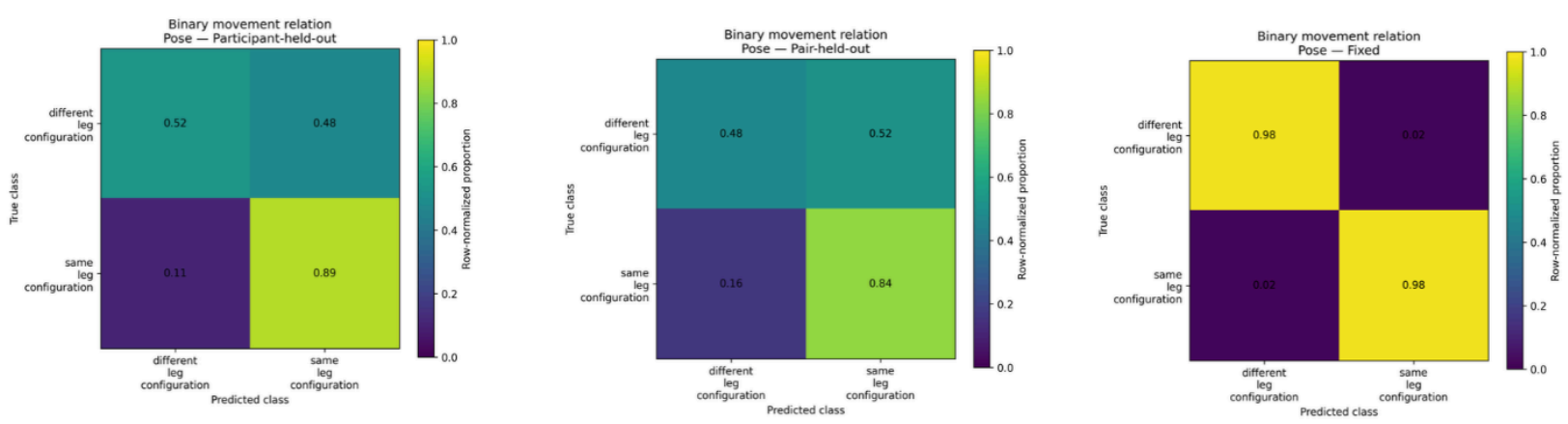

**Figure 8. Pose confusion matrices for binary movement-relation classification.** Row-normalized confusion matrices under duplicate-safe fixed, pair-held-out, and participant-held-out evaluation. The held-out protocols show a pronounced reduction in recall for different_leg_configuration, while same_leg_configuration remains comparatively well recognized.

### 8.6 Behaviour of RGB–Pose Fusion

Late fusion produces the highest mean performance on both duplicate-safe fixed benchmarks and a modest improvement over pose for four-class pair-held-out evaluation. Its benefits are not consistent across the held-out settings.

For binary pair-held-out evaluation, fusion improves over pose in one of the four outer folds, ties pose in two folds, and performs below pose in one. The mean difference relative to pose is approximately −0.002 macro-F1.

For four-class pair-held-out evaluation, fusion improves the overall mean by approximately 0.014 macro-F1 relative to pose, although the improvement occurs in only one of the four outer folds.

The clearest fusion failure occurs in the binary participant-held-out diagnostic. Across the eight eligible participants, fusion improves over pose for two participants, ties for two, and performs worse for four. Its overall mean is approximately 0.100 macro-F1 below pose.

Post-benchmark diagnostic analysis indicates that this behaviour is associated with validation-based selection of the RGB fusion weight.

For participant P03, pose obtains approximately 0.917 macro-F1 when averaged across the three training seeds, while RGB obtains approximately 0.353. Despite this difference on the held-out participant, validation selects

$\alpha = 1.0$

for all three runs. The fusion model therefore reduces to RGB-only prediction and obtains approximately 0.353 macro-F1.

The validation advantage leading to these selections is comparatively small. Across the three P03 runs, the RGB-heavy solution improves validation macro-F1 over pose by approximately 0.026–0.039, while held-out participant performance decreases by approximately 0.538–0.608 macro-F1.

A related case occurs for P11 under seed 43, where validation again selects $\alpha = 1.0$,

and test macro-F1 decreases by approximately 0.651 relative to pose.

Across the binary participant-held-out runs, the descriptive correlation between the selected RGB weight and the test-set fusion improvement relative to pose is approximately

$r = -0.735$.

Given the small number of participant-level evaluations, this correlation is reported descriptively rather than as a population-level statistical estimate.

These analyses were conducted after completion of the locked benchmark. Test results were not used to revise the fusion weights or alter the Baseline V1 predictions.

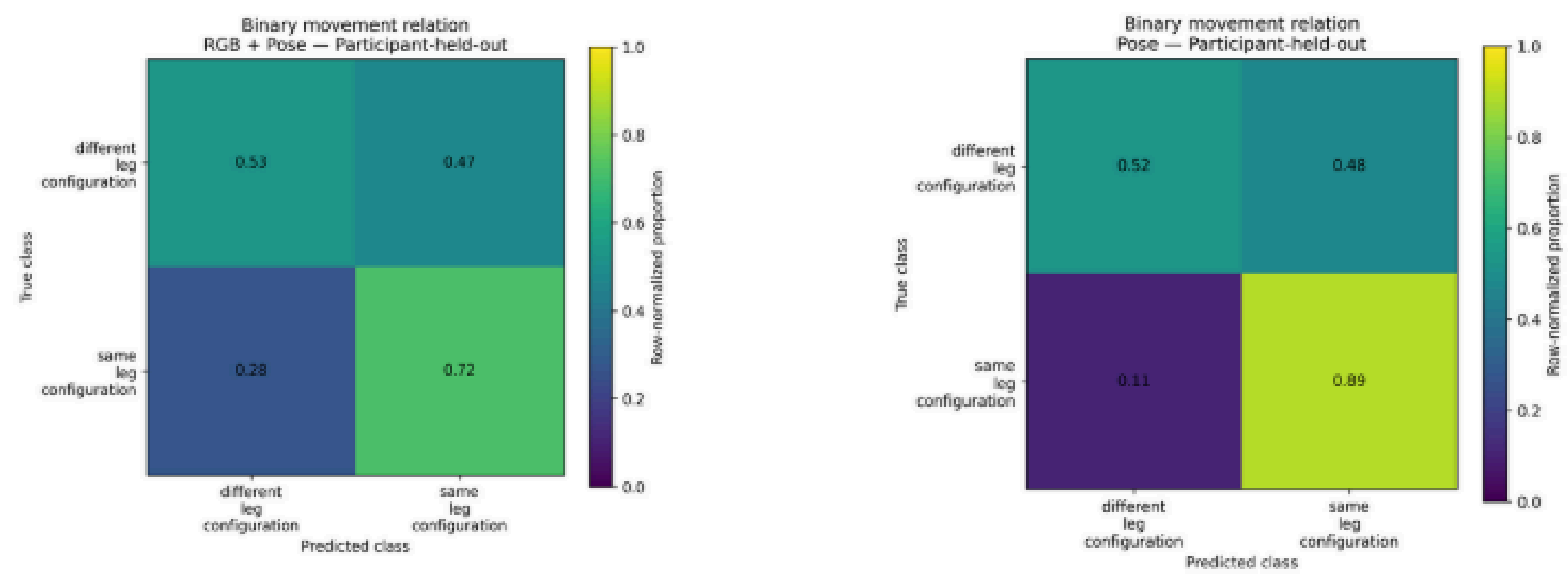

**Figure 9. Binary participant-held-out confusion matrices for pose and RGB + pose fusion.** Pose retains comparatively high recall for the same-leg relation, whereas fusion substantially reduces same-leg recall under the participant-held-out diagnostic. The comparison illustrates the consequences of unstable validation-selected RGB weighting in some participant folds.

## 8.7 Outcome-Level Evaluation

Baseline V1 also reports rule-derived outcome accuracy as an auxiliary metric.

Because this outcome is computed deterministically from the movement prediction and the encoded benchmark rule, it is not treated as an independently annotated visual-recognition target.

For four-class recognition on the duplicate-safe fixed benchmark, outcome accuracy is 0.979 for RGB, 0.962 for pose, and 0.986 for fusion.

Under pair-held-out evaluation, the corresponding outcome accuracies are 0.716 for RGB, 0.833 for pose, and 0.854 for fusion.

Under participant-held-out evaluation, outcome accuracy is 0.636 for RGB, 0.863 for pose, and 0.864 for fusion.

Outcome accuracy can remain higher than four-class movement-classification performance because some errors between four-class configurations preserve the same binary movement relation. For example, confusing one same-leg configuration with the other changes the fine-grained movement label without necessarily changing the rule-derived interpretation of the interaction.

For the binary task, outcome accuracy is equivalent to binary classification accuracy under the encoded benchmark rule and therefore does not constitute an additional independent metric.

### 8.8 Summary of Empirical Findings

The Baseline V1 experiments establish four principal observations.

First, all learned representations perform strongly on the duplicate-safe fixed benchmark. Four-class macro-F1 ranges from 0.955 to 0.976, while binary macro-F1 ranges from 0.932 to 0.989.

Second, these fixed-benchmark results do not transfer uniformly to held-out participant structures. The largest example is four-class RGB recognition, which decreases from 0.955 macro-F1 on the fixed benchmark to 0.418 under pair-held-out evaluation and 0.384 under participant-held-out evaluation.

Third, pose is substantially more stable than frozen RGB for four-class recognition under both forms of held-out evaluation, retaining 0.811 macro-F1 for unseen participant pairings and 0.839 for unseen participants.

Fourth, RGB-pose late fusion is not uniformly beneficial. It produces the highest fixed-benchmark results and a modest improvement for four-class pair-held-out evaluation, but does not outperform pose under four-class participant-held-out evaluation and performs substantially below pose for binary participant-held-out evaluation.

Taken together, these results show that performance on the duplicate-safe fixed benchmark alone substantially understates the generalization difficulty exposed when participant pairings or participant identities are held out.

# 9. Discussion

The Ampe-DB experiments show that benchmark performance depends strongly on how generalization is defined. The duplicate-safe fixed benchmark suggests that the movement-recognition tasks can be handled effectively by relatively simple reference models. The held-out protocols reveal a different picture, particularly for frozen RGB representations.

Three findings are central. First, high performance under a fixed split does not imply equally strong transfer to unfamiliar participant pairings or participants. Second, pose representations are substantially more stable than frozen RGB features for four-class recognition under both held-out settings. Third, simple RGB-pose late fusion is beneficial in some conditions but can fail when validation-based modality weighting does not transfer to the held-out distribution.

## 9.1 Fixed-Split Performance Does Not Guarantee Generalization

The duplicate-safe fixed benchmark produces high performance across all learned baselines. Because exact audiovisual duplicates are already prevented from crossing partitions, these results are not explained by direct duplicate leakage.

However, the held-out evaluations show that exact-content separation alone does not guarantee transfer across people or interaction structure.

The sharp decline of the four-class RGB baseline under pair-held-out and participant-held-out evaluation indicates that information supporting strong prediction within the fixed benchmark does not transfer equally well when familiar participant relationships are removed.

RGB representations preserve considerably more than the target movement. Clothing, body appearance, background, illumination, camera characteristics, and other visual context remain available to the model alongside action-related information. Some of these signals may be associated with the development distribution and may therefore contribute to prediction without transferring reliably to held-out people or pairings.

The present experiments do not isolate which visual factors account for the observed RGB generalization gap. They therefore do not establish that the model explicitly memorizes participant identity or any specific contextual feature.

What the experiments do establish is that the frozen RGB representation transfers substantially less reliably than the pose representation under the participant-aware evaluation protocols.

This supports the use of pair-aware and participant-aware evaluation for human-activity datasets containing recurring performers or recurring interaction structures.

### 9.2 Pose Is Better Matched to the Four-Class Ampe Task

The strongest held-out results in the four-class task come from the pose representation.

This is consistent with the structure of the recognition problem. The target categories are defined primarily by the body configurations of two participants, particularly the left-right relationship between their leg movements. A representation centred on body landmarks therefore preserves information that is closely aligned with the target while reducing much of the appearance and scene information retained by RGB.

The result should not be interpreted as a general claim that pose is superior to RGB for human activity recognition. Pose representations omit information that can be important in other tasks, including objects, facial expression, hand appearance, environmental context, and other visual cues.

Pose extraction is also itself derived from RGB observations and can introduce estimation error.

For Ampe-DB, however, the held-out results indicate that explicit body-configuration information is particularly well aligned with the current four-class recognition task.

The strength of the compact BiGRU baseline also suggests that the pose modality has substantial remaining modelling potential. More specialized architectures, including spatial-temporal graph models and transformer-based approaches, could explicitly represent anatomical structure, cross-player relationships, and temporal dependencies beyond those captured by the current baseline.

### 9.3 The Binary Task Exposes a Distinct Generalization Challenge

Reducing the four movement configurations to two relational categories might be expected to simplify recognition substantially. The results show that this is not sufficient to remove the held-out generalization problem.

The confusion-matrix analysis indicates that much of the difficulty is concentrated in the different_leg_configuration class.

The corresponding four-class results show a similar pattern. The two cross-leg configurations, left_right and right_left, lose more recall under held-out evaluation than the same-leg configurations.

This suggests that the binary task is not difficult merely because of class imbalance or output-space size. Instead, the cross-leg relationships themselves appear less consistently recognized across unfamiliar people and pairings.

Several factors could contribute to this pattern, including variation in body orientation, overlap between participants' legs, differences in movement timing, landmark visibility, and pose-estimation quality.

These possibilities remain hypotheses. Baseline V1 does not independently measure their contributions.

A targeted follow-up analysis could therefore examine landmark confidence, temporal offsets between participants, body orientation, occlusion, and leg visibility to determine which factors are most strongly associated with cross-leg errors.

### 9.4 Pair-Held-Out and Participant-Held-Out Evaluation Capture Different Shifts

The two held-out protocols provide complementary rather than interchangeable measurements.

Pair-held-out evaluation removes a specific participant combination from model development while allowing an individual participant to appear elsewhere with another partner.

Participant-held-out evaluation removes the target participant entirely from training and validation.

These protocols therefore alter different aspects of the interaction distribution. Their results should not be interpreted as points on a simple easy-to-hard scale.

This distinction is particularly important because the participant-held-out diagnostic is restricted to participants with complete target-class coverage, whereas the pair-held-out benchmark operates over a different set of evaluation units.

Consequently, the similar four-class pose performance observed across these settings should not be interpreted as evidence that participant-held-out recognition is inherently easier than pair-held-out recognition.

The more useful conclusion is that pose remains comparatively stable under both forms of shift, while the frozen RGB representation performs substantially worse in both.

The pair-held-out benchmark should therefore remain the principal generalization benchmark, with participant-held-out evaluation serving as a complementary diagnostic of transfer to unseen performers.

### 9.5 Simple Multimodal Fusion Is Not Reliably Beneficial

The fusion results show that complementary modalities do not automatically produce a stronger model.

RGB-pose late fusion improves the fixed-benchmark results and produces a modest gain over pose for four-class pair-held-out evaluation. However, it does not improve over pose in the four-class participant-held-out diagnostic and performs substantially below pose for binary participant-held-out evaluation.

The post-benchmark analysis suggests that the problem lies partly in modality selection rather than in an inherent incompatibility between RGB and pose.

Fusion weights are selected using validation macro-F1. In some participant-held-out runs, relatively small validation advantages lead the procedure to assign large or complete weight to RGB even though RGB transfers poorly to the subsequently held-out participant.

The P03 and P11 cases illustrate this failure mode particularly clearly.

These observations indicate that validation performance can provide a misleading basis for modality weighting when the validation distribution does not reproduce the type of identity shift present at test time.

The descriptive negative association between selected RGB weight and held-out fusion improvement is consistent with this interpretation, although the number of participant-level evaluations is too small to support a broader statistical claim.

Future multimodal approaches should therefore consider model-selection procedures that better match the intended generalization setting. Possibilities include nested participant-aware validation, group-aware modality calibration, uncertainty-based weighting, feature-level multimodal learning, and jointly trained models evaluated using identity-separated validation.

Such approaches would require new experiments and should not be used retrospectively to modify the frozen Baseline V1 results.

### 9.6 Implications for Benchmark Design

One of the broader findings from Ampe-DB is methodological rather than architectural.

A benchmark can appear close to saturated under a fixed split while still containing a substantial unresolved generalization problem.

In Ampe-DB, RGB and pose are nearly indistinguishable on the fixed four-class benchmark, yet their behaviour diverges sharply under pair-held-out and participant-held-out evaluation.

That difference would not be visible if the benchmark reported only one predefined split.

Evaluation design is therefore part of the contribution of Ampe-DB rather than merely an implementation detail.

The combination of duplicate-aware partitioning, pair-held-out evaluation, and participant-held-out diagnostics separates different sources of dependence that could otherwise be hidden inside a conventional train-test split.

Future work using Ampe-DB should report results under the protocol appropriate to the research question and should not compare scores from different protocols as though they estimate the same quantity.

### 9.7 Cultural Context as a Source of Technical Structure

Ampe-DB was motivated partly by the limited representation of culturally situated African activities in widely used human-action datasets. The benchmark also shows that cultural context and technical research questions need not be treated as separate objectives.

The structure of Ampe directly generates the computational problem.

The game involves coordinated movement between two people, fine-grained left-right distinctions, relational same-leg and different-leg states, and temporal interaction.

These properties naturally motivate questions in fine-grained activity recognition, human-human interaction analysis, pose modelling, multimodal learning, and generalization across people.

Ampe-DB therefore contributes more than geographic representation to an existing action category. The activity itself determines the structure of the labels, the interaction representation, and the evaluation questions.

At the same time, the benchmark abstraction should not replace the cultural context from which it is derived.

Ampe is a living Ghanaian game and social practice. The four computational movement classes used in Ampe-DB represent only the aspects of the activity required for the current recognition tasks.

Dataset documentation and downstream research should therefore distinguish clearly between computational analysis of the benchmark and broader claims about Ampe as a cultural practice.

## 9.8 Opportunities for Future Modelling

The Baseline V1 results leave several modelling directions open.

More expressive pose architectures could model anatomical and cross-player structure directly. Spatial-temporal graph networks could represent relationships among joints and between participants, while transformer-based models could examine longer-range temporal dependencies and interactions between corresponding body regions.

Future RGB experiments could evaluate whether end-to-end fine-tuning, stronger contemporary video encoders, spatial attention, or representations designed to reduce sensitivity to performer appearance improve held-out generalization.

Multimodal models could also move beyond scalar late fusion. Feature-level interaction, cross-modal attention, or participant-aware calibration may provide more reliable integration of RGB and pose.

The retained audio stream represents an additional potential modality. Rhythm, vocalization, foot contact, and other acoustic information may contain useful interaction cues. However, any future audio benchmark should first establish what information is present and whether the modality introduces participant-specific shortcuts.

Future versions of Ampe-DB could also expand the number of participants, partner combinations, viewpoints, recording environments, and gameplay styles. Such extensions would allow stronger evaluation of whether models capture the underlying movement structure across broader real-world variation.

## 9.9 Overall Interpretation

The central finding of Baseline V1 is not simply that one modality obtains a higher score than another.

The more important result is that the apparent difficulty of Ampe recognition changes substantially when the evaluation protocol changes.

Under the duplicate-safe fixed benchmark, relatively simple models achieve strong performance. When participant pairings or participant identities are held out, substantially more difficult generalization behaviour becomes visible.

Pose retains much more of its four-class performance under these shifts than the frozen RGB representation. Simple late fusion can improve performance in some settings, but can also inherit the weaker modality's generalization failure when validation-based weighting does not transfer.

Together, these findings position Ampe-DB as a benchmark in which movement representation, interaction structure, participant shift, and evaluation design are closely connected. They also establish clear directions for future modelling without treating the Baseline V1 systems as final solutions.

# 10. Limitations and Ethical Considerations

## 10.1 Dataset Scale and Participant Coverage

Ampe-DB V1 contains 637 canonical interaction records involving 15 participants and 11 unordered participant pairs. This scale is sufficient to establish the benchmark and reveal substantial differences between fixed and held-out evaluation, but it remains small relative to large-scale human-action datasets.

The limited participant population constrains the conclusions that can be drawn about population-level generalization. Performance under the current participant-held-out diagnostic should therefore not be interpreted as evidence that a model will generalize equally well to the broader population of Ampe players.

Participant-level class coverage is also uneven. Six participants contain all four movement configurations required for the four-class participant-held-out diagnostic, while eight contain both target classes required for binary participant-held-out evaluation. Consequently, participant-held-out results describe the eligible subset rather than all 15 participants.

This limitation is reflected directly in the benchmark design: participant-held-out evaluation is treated as a diagnostic rather than the principal generalization benchmark.

Future versions of Ampe-DB would benefit from larger participant coverage, more complete movement-class representation per participant, and a wider range of participant pairings.

### 10.2 Limited Participant-Pair Structure

The current dataset contains 11 unordered participant pairs.

This restricts the range of interpersonal variation represented in Baseline V1. Individual participants may also appear in more than one pair, meaning that pair-level and participant-level variation are not statistically independent.

The four-fold pair-held-out benchmark prevents test pairs from appearing in model-development data, but it cannot reproduce the diversity of interaction structure that would be available in a substantially larger and more densely connected participant graph.

Future data collection should therefore increase both the number of participants and the number of partner combinations. A larger interaction graph would support stronger analysis of transfer across partners, movement styles, and interaction dynamics.

### 10.3 Recording Conditions and Viewpoint

Ampe-DB was collected under a limited range of recording conditions and does not represent the full variety of environments in which Ampe may be played.

Real-world gameplay may differ in lighting, camera viewpoint, distance, background, clothing, playing surface, crowd density, and occlusion.

The predominantly frontal, full-body recording design is particularly relevant. It makes the leg configurations required by the benchmark comparatively observable, but it does not establish viewpoint invariance.

Models trained on Ampe-DB V1 may therefore behave differently when applied to footage captured under substantially different visual conditions.

Future versions should include broader environmental and viewpoint variation while maintaining sufficient visibility to support the fine-grained left/right movement labels.

### 10.4 Pose as a Derived Representation

The pose modality is automatically estimated from RGB video rather than obtained from direct motion capture.

It therefore inherits limitations from the person-detection and pose-estimation pipeline. Potential sources of error include occlusion between participants, rapid movement, motion blur, overlapping limbs, missed detections, unusual body configurations, and uncertainty in left/right landmark localization.

These limitations are particularly relevant to Ampe because the benchmark depends on fine-grained relationships between the participants' legs.

Baseline V1 preserves source-timeline alignment and uses explicit player-presence indicators when pose observations are unavailable. This procedure prevents missing detections from implicitly compressing the temporal sequence, but it does not correct landmarks that are detected but inaccurately localized.

The strong held-out performance of the pose model should therefore not be interpreted as evidence that the extracted skeletons constitute perfect measurements of the underlying movements.

Future work could include pose-confidence analysis, manually reviewed quality subsets, or comparisons among multiple pose-estimation systems.

### 10.5 Exact Duplicate Content

The 637 canonical records correspond to 557 unique audiovisual SHA-256 hashes. Some canonical records therefore contain byte-identical audiovisual content.

These records are retained because canonical record identity and audiovisual-content identity represent different properties of the dataset. Their presence nevertheless creates a methodological risk if future users construct evaluation splits without respecting duplicate-content groups.

Baseline V1 controls this risk by constraining identical-content records to the same train, validation, or test partition.

Researchers constructing alternative protocols should maintain equivalent content-level separation. Otherwise, exact-content leakage could inflate evaluation performance.

### 10.6 Scope of the Baseline Models

The models reported in this paper are intentionally limited reference baselines.

The RGB backbone is frozen rather than fine-tuned end-to-end, the pose representation is modelled using a compact bidirectional GRU, and multimodal integration is implemented using scalar probability-level late fusion.

The reported results should therefore not be interpreted as estimates of the maximum achievable performance on Ampe-DB.

Likewise, the weak held-out performance of the frozen RGB representation should not be generalized to all RGB-based action-recognition systems. End-to-end adaptation, stronger video encoders, alternative spatial-temporal representations, and other training strategies may behave differently.

Baseline V1 is intended to establish reproducible reference behaviour and expose the generalization properties of the benchmark.

### 10.7 Limits of the Fusion Analysis

The fusion diagnostic is based on a relatively small number of participant and pair evaluation units.

The negative association observed between selected RGB weight and participant-held-out fusion improvement is useful for understanding the behaviour of the Baseline V1 selection procedure, but it should not be treated as a population-level statistical relationship.

The analysis was also conducted after the benchmark experiments had been completed. It was therefore used only to interpret the frozen results and not to modify fusion weights or replace test predictions.

Claims about more robust fusion strategies will require new experiments in which the model-selection procedure is specified before test evaluation.

### 10.8 Cultural Scope and Interpretation

Ampe is a living Ghanaian game and may vary in rules, terminology, rhythm, playing style, participant roles, and local interpretation.

Ampe-DB necessarily represents the form of the activity captured during the present data-collection process. It should not be interpreted as a complete computational representation of every way in which Ampe is played or understood across Ghana or elsewhere.

The four movement configurations used by the benchmark are analytical representations of observed leg states. They are useful for machine-learning evaluation but do not capture the full social, cultural, rhythmic, competitive, and interpersonal dimensions of the game.

Researchers should therefore distinguish conclusions about the Ampe-DB benchmark from broader claims about Ampe as a cultural practice.

A model that correctly classifies a movement configuration should not be interpreted as possessing comprehensive understanding of the game or the cultural knowledge associated with it.

### 10.9 Participant Privacy

The RGB recordings contain identifiable participant information, including faces, bodies, clothing, voices, and potentially other contextual features.

Raw RGB video and synchronized audio therefore present greater privacy risk than normalized pose representations.

Research questions that do not require identifiable appearance information may be supported using derived pose representations and pseudonymous metadata. Access to identifiable source media should follow the permissions established by the project's consent and data-governance documentation.

Audio is retained in the canonical recordings but is not used in Baseline V1. Because voices can also be identifying, audio should be treated as participant-identifiable data rather than as an automatically unrestricted modality.

### 10.10 Consent, Research Ethics, and Data Governance

Participants provided informed consent for the collection and research use of Ampe-DB recordings, including the public distribution of the participant-derived materials released as part of the dataset. The project team confirms that the necessary participant consent was obtained. Consent records are maintained securely by the data stewards and are not included in the public dataset to protect participant privacy.

Ampe-DB includes RGB recordings with embedded audio, normalized pose sequences, pose-derived skeleton outputs, and pseudonymous participant-linked metadata. The released metadata use stable pseudonymous identifiers (P01–P15). Real participant names, contact details, internal identity mappings, and identity-review materials are excluded from the publication.

The dataset is hosted through the official Ampe-DB Hugging Face repository under a custom Research Use License Notice and Research Data Use Agreement. Direct repository downloads do not require case-by-case research approval or completion of the separate declaration form intended for the project's future website. Users remain subject to the documented restrictions concerning redistribution, participant re-identification, biometric identification, surveillance, profiling, commercial use, and cultural responsibility.

The project team confirms that participant consent for external distribution of the released materials has been secured. The relevant documentation is retained by the data stewards, who remain responsible for maintaining consent records and ensuring that future dataset releases and uses respect the permissions obtained.

The available documentation does not establish a verified institutional ethics-review approval or exemption identifier. Accordingly, this manuscript does not assert institutional approval or exemption unsupported by the available records.

Because Ampe is a Ghanaian cultural practice, responsible dataset governance also requires preserving the cultural context of the game and avoiding exploitative, stereotypical, or decontextualized representations.

### 10.11 Responsible Use

Ampe-DB is intended for research and educational work involving human activity recognition, interaction analysis, pose modelling, multimodal learning, and related computational questions.

Because the dataset contains or derives from identifiable human behaviour, applications involving surveillance, biometric identification, harmful participant profiling, or inference of sensitive personal characteristics fall outside the intended benchmark use and require separate ethical consideration.

Participant identifiers should be interpreted only as pseudonymous dataset-level identifiers used for benchmark construction. They should not be treated as semantic descriptions of the individuals.

Researchers should also avoid using model predictions to make unsupported claims about participants' abilities, behaviour, identity, or cultural knowledge.

### 10.12 Summary of Limitations

Ampe-DB provides a structured and culturally grounded benchmark for studying two-person movement, but its current scope remains deliberately bounded.

Its principal limitations include the small participant population, restricted participant-pair graph, incomplete participant-level class coverage, limited recording environments and viewpoints, reliance on automatically estimated 2D pose, the presence of exact duplicate records requiring leakage-aware handling, and the identifiable nature of the original RGB and audio recordings.

The baseline architectures are also intentionally limited, and the multimodal fusion analysis is based on a small number of held-out evaluation units.

These constraints define the conditions under which the current results should be interpreted and identify concrete priorities for future versions of the dataset.

## 11. Conclusion

We introduced Ampe-DB, a multimodal dataset and benchmark for human activity and interaction analysis based on the Ghanaian game Ampe. Ampe-DB V1 contains 637 canonical two-person interaction records involving 15 participants and provides RGB video, synchronized audio, normalized two-person pose representations, participant and pair metadata, four-class movement labels, and deterministically derived binary movement-relation labels.

A central contribution of the benchmark is its evaluation design. In addition to a duplicate-safe fixed split, Ampe-DB includes pair-held-out evaluation for transfer to unseen participant pairings and a participant-held-out diagnostic for transfer when a performer has been excluded from model development.

The Baseline V1 experiments show why this distinction matters. RGB, pose, and RGB-pose fusion all perform strongly on the duplicate-safe fixed benchmark, but their behaviour differs substantially under held-out evaluation. For four-class recognition, pose retains considerably stronger performance than the frozen RGB representation on unseen participant pairs and unseen participants. Simple probability-level fusion improves performance in some settings but is not consistently beneficial, particularly when validation-selected modality weights fail to transfer under participant shift.

These findings demonstrate that fixed-split performance alone does not characterize the full difficulty of Ampe-DB. The benchmark instead exposes a generalization problem involving movement representation, participant identity, and interaction structure.

More broadly, Ampe-DB shows how a culturally situated activity can generate technically meaningful research questions in fine-grained activity recognition, human-human interaction analysis, pose modelling, multimodal learning, and generalization across people.

The current release should be viewed as a starting point rather than a complete computational representation of Ampe. Future work can extend participant and partner diversity, recording conditions, viewpoints, modalities, and model architectures while preserving the leakage-aware and participant-aware evaluation principles established in Baseline V1.

## 12. Data and Code Availability

Ampe-DB V1 (version 1.0.0) is available through the official Hugging Face dataset repository:

https://huggingface.co/datasets/RuthBiney/Ampe-DB

The release contains 637 canonical interaction records and 3,185 primary data files, comprising RGB videos with synchronized embedded audio, normalized two-person pose CSV files, and pose-derived skeleton visualizations and keypoint representations in MP4, CSV, and JSON formats.

Supporting materials include verified movement annotations, pseudonymous participant and pair metadata, the canonical manifest, schema and data dictionary, predefined duplicate-safe and participant-aware benchmark splits, technical documentation, citation metadata, and research-use conditions.

A master SHA-256 register covering all 3,185 primary published files is available at:

https://huggingface.co/datasets/RuthBiney/Ampe-DB/tree/main/checksums

The dataset is distributed under the Ampe-DB Research Use License Notice, Research Data Use Agreement, and associated Access Terms. Direct Hugging Face downloads do not require case-by-case manual approval or completion of the separate researcher-use declaration intended for the project's future website. Access does not authorize unrestricted redistribution or prohibited participant-directed uses. Distribution and downstream use remain subject to applicable participant rights and permissions.

The official benchmark splits preserve exact-content duplicate separation and support the duplicate-safe fixed, four-fold pair-held-out, and participant-held-out diagnostic protocols reported in this paper.

Baseline implementation code, evaluation scripts, trained-model outputs, and detailed experimental artifacts are maintained separately from the core dataset release. Their public availability should not be inferred from the publication of the dataset repository. The manuscript documents the model architectures, preprocessing, training seeds, benchmark protocols, and evaluation procedures used in Baseline V1.

Researchers should cite the dataset using its accompanying CITATION.cff metadata and record the release version and repository revision used for their experiments.